\pdfoutput=1
\documentclass[11pt]{article}
\usepackage[final]{acl}

\usepackage{times}
\usepackage{latexsym}
\usepackage[T1]{fontenc}
\usepackage[utf8]{inputenc}
\usepackage{microtype}
\usepackage{inconsolata}

\usepackage{graphicx}
\usepackage{xcolor}
\usepackage{colortbl}

\usepackage{amsmath}
\usepackage{courier}
\usepackage{amsfonts}
\usepackage{array}
\usepackage{subfig}
\usepackage{CJKutf8}
\usepackage{arydshln}
\usepackage{caption}
\usepackage{hyperref}
\usepackage{algorithm}
\usepackage{algorithmic}
\usepackage{adjustbox}
\usepackage{float}
\usepackage{booktabs}

\usepackage{amsmath}
\usepackage{newfloat}
\usepackage{listings}
\usepackage{wrapfig}        
\usepackage{url}
\usepackage{amsfonts}       
\usepackage{nicefrac}       
\usepackage{graphicx}
\usepackage{amsmath}
\usepackage{multirow} 
\usepackage{paralist} 
\usepackage{subcaption}
\usepackage{enumitem}
\usepackage[most]{tcolorbox}
\usepackage{listings}
\lstdefinestyle{mystyle}{
    basicstyle=\ttfamily\small, 
    keywordstyle=\color{blue},  
    commentstyle=\color{gray},  
    stringstyle=\color{red},
    backgroundcolor=\color{lightgray!20},
    numberstyle=\tiny\color{gray},
    breaklines=true,
    frame=single,
    rulecolor=\color{black!50}
}
\definecolor{egggreen}{HTML}{d7e6e8}
\definecolor{lightskyblue3}{HTML}{dbdbeb}
\definecolor{lightyellowgreen3}{HTML}{d6ecf0}
\definecolor{constraint}{HTML}{eee2f0}
\definecolor{LemonChiffon2}{HTML}{EEE9BF}
\definecolor{Ivory2}{HTML}{EEEEE0}
\definecolor{LavenderBlush2}{HTML}{EEE0E5}
\definecolor{RED}{HTML}{FCBBC3}
\definecolor{maroon}{cmyk}{0,0.87,0.68,0.32}

\newtcolorbox{mybox}[1][]{
	width=\columnwidth,
	colback = gray!6, 
	colframe = black, 
	boxrule = 0.8pt,
	boxsep=0pt,left=10pt,right=10pt,top=8pt,bottom=8pt,
	fontupper=\linespread{1.1}\selectfont,
	title=#1}

\definecolor{myPurple}{rgb}{0.4706, 0, 0.3922}
\definecolor{tkcolor}{RGB}{224,223,255}
\newtcolorbox{takeaways}[1][]{
	width=\columnwidth,
	colback = tkcolor, 
	colframe = tkcolor, 
	boxsep=0pt,left=10pt,right=10pt,top=2pt,bottom=2pt,
	fontupper=\linespread{0.9}\selectfont,
	title=#1}
\title{What are the Essential Factors in Crafting Effective Long Context Multi-Hop Instruction Datasets? Insights and Best Practices}

\author{Zhi Chen$^{\clubsuit}$\thanks{\ \ Equal Contribution} \ \ Qiguang Chen$^{\heartsuit}$\footnotemark[1] \ \  Libo Qin$^\diamondsuit$\thanks{\ \ Corresponding Author} \ \  Qipeng Guo$^\clubsuit$ \ \  \textbf{Haijun Lv}$^\clubsuit$ 
\AND {}\vspace{-0.7cm}\\ \textbf{Yicheng Zou}$^\clubsuit$ \ \ \textbf{Hang Yan} $^\clubsuit$ \ \  \textbf{Kai Chen} $^\clubsuit$ \ \  \textbf{Dahua Lin}$^\clubsuit$\\
$^\clubsuit$ Shanghai Artificial Intelligence Laboratory \\
$^\ddagger$ Research Center for Social Computing and Interactive Robotics \\
$^\ddagger$ Harbin Institute of Technology\\
$^\diamondsuit$ School of Computer Science and Engineering, Central South University\\[1ex]
\texttt{chenzhi@pjlab.org.cn, qgchen@ir.hit.edu.cn}}

\begin{document}
\maketitle

\begin{abstract}
Recent advancements in large language models (LLMs) with extended context windows have significantly improved various tasks. To improve long-context capabilities, much work focuses on augmenting LLM's capabilities with synthetic data. Existing methods often leverage the Self-Instruct framework to generate long-context instruction-tuning data. However, our preliminary experiments show that fewer than 35\% of samples generated by Qwen-2$_{72B}$ are multi-hop, and over 40\% exhibit poor quality, limiting comprehensive understanding and further research.
To address this, we propose the Multi-agent Interactive Multi-hop Generation (\texttt{MIMG}) framework, which integrates a quality verification agent, a single-hop question generation agent, a multiple question sampling strategy, and a multi-hop question merger agent. This framework significantly improves data quality, with high-quality, multi-hop, and diverse data. Furthermore, we conduct a thorough analysis of document selection, question merging, and validation techniques through extensive experiments across various models. Our results demonstrate that synthetic high-quality long-context instruction data can enhance model performance, surpassing even models trained on larger amounts of human-annotated data. Our code and relevant data are available at: \url{https://github.com/WowCZ/LongMIT}.
\end{abstract}

\section{Introduction}
\begin{figure}[t]
	\centering
	\includegraphics[width=0.48\textwidth]{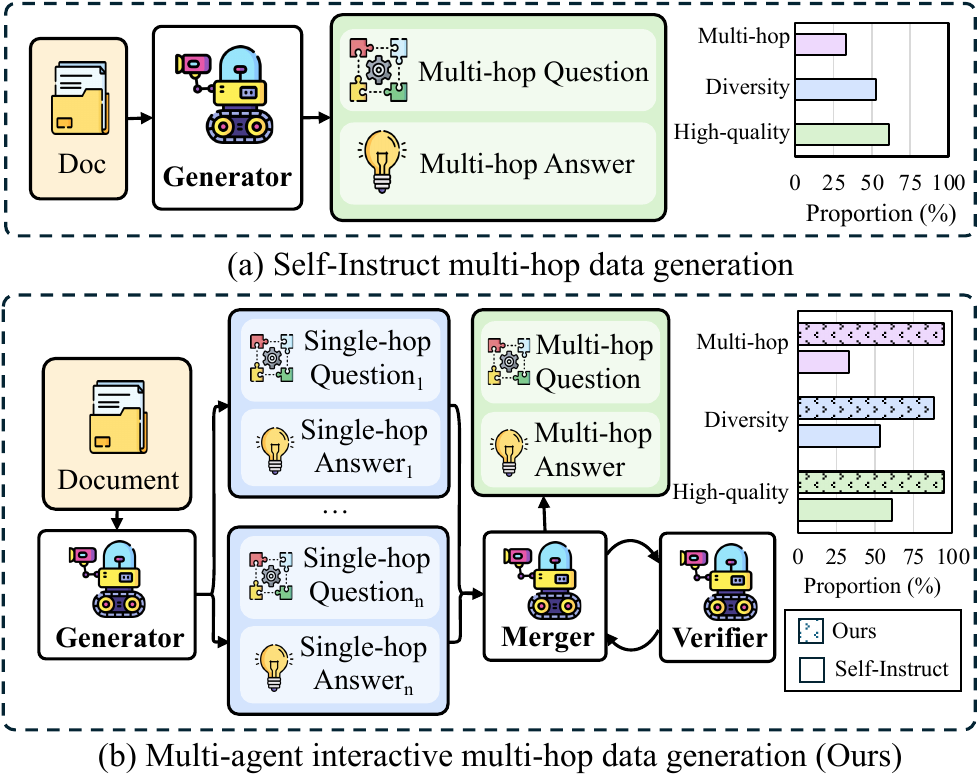}
	\caption{Comparison between traditional self-instruct-based data synthesis method and our Multi-agent Interactive Multi-hop Generation (\texttt{MIMG}) framework, where all data are generated by Qwen-2$_{72B}$~\citep{yang2024qwen2}.
	}
	\label{fig:intro}
\end{figure}
\begin{figure*}[t]
	\centering
	\includegraphics[width=0.98\textwidth]{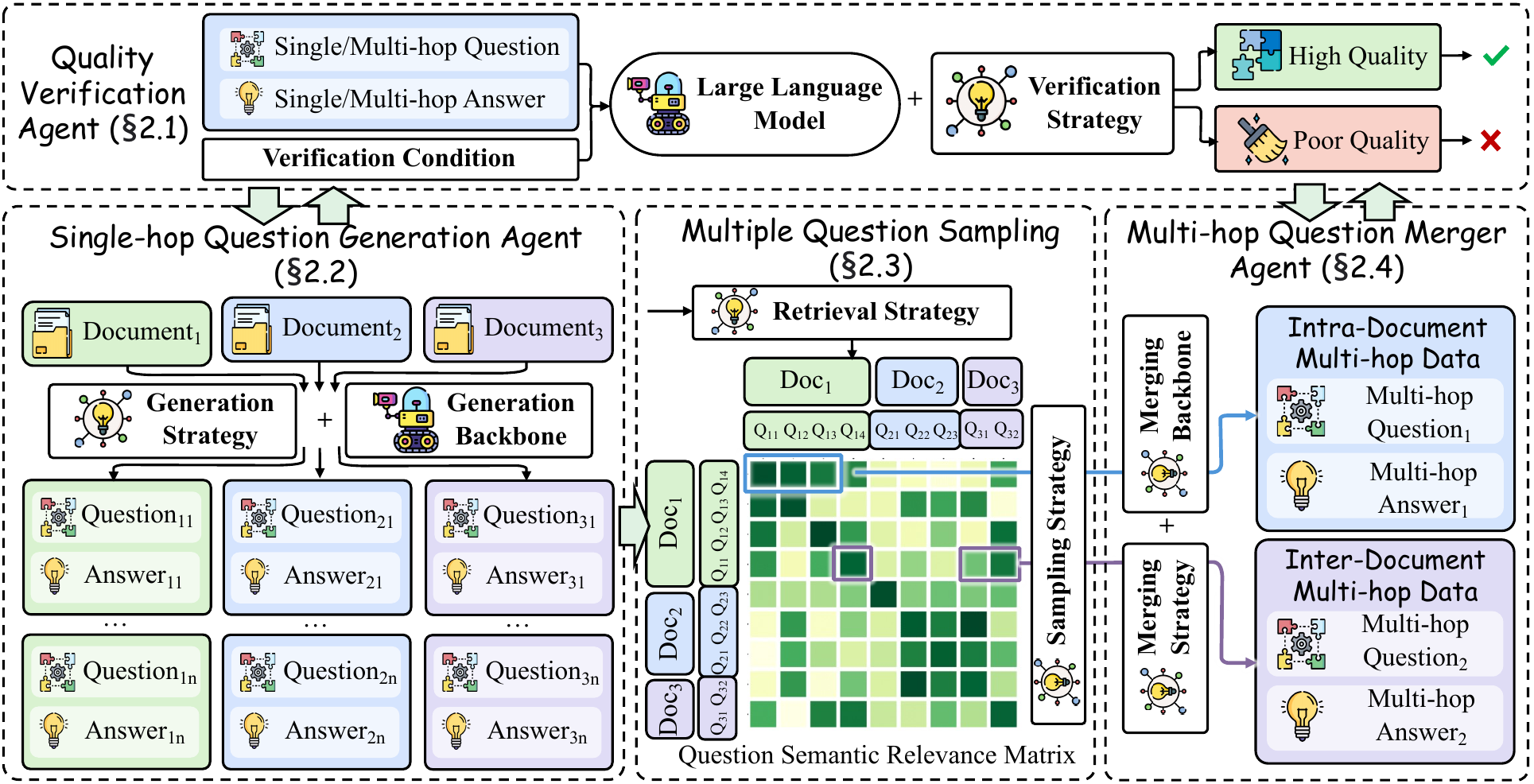}
	\caption{The overall process of Multi-agent Interactive Multi-hop Generation (\texttt{MIMG}) data synthesis framework.}
	\label{fig:framework}
\end{figure*}

Recently, large language models (LLMs) with long-context windows have significantly improved tasks such as information extraction, question answering, and even complex planning scenarios~\citep{liu2024lost,bai2024longbench,hu2025text2world,xu2024concise}. Research on developing long-context LLMs has predominantly focused on extending the context window~\citep{ding2024longrope,jin2024llm,peng2024yarn}. Nevertheless, in practical applications, merely expanding the context window is insufficient for effectively utilizing long-context~\citep{hsieh2024ruler,huang2024well}, which presses a need for training to optimize utilization of long-context~\citep{zhang2024extending}, especially in instruction-tuning (IT)~\citep{fu2024data}.
In the IT phase, a large amount of high-quality long-context IT data is required. However, acquiring such data is challenging, with annotation costs significantly higher than those for short-context data~\citep{bai2024longbench,xiong-etal-2024-effective}. To address this, \citet{xiong2023effective} and \citet{bai2024long} have explored leveraging LLMs to generate IT data using the Self-Instruct framework~\citep{wang-etal-2023-self-instruct}, thereby mitigating the scarcity of long-context IT data.

Moreover, the challenge often lies not in extracting single-hop information, but in integrating multiple hops of data from long contexts to derive complex conclusions.
However, existing studies struggle to generate high-quality, multi-hop IT data, primarily due to insufficient focus on the data synthesis process and the factors influencing data effectiveness. As illustrated in Figure~\ref{fig:intro} (a), our preliminary manual annotation experiments show that direct self-instruction yields less than 35\% multi-hop samples, with high-quality examples representing only 60\%. Additionally, sample diversity remains problematic, with over 45\% of the samples exhibiting semantic duplication. These issues hinder comprehensive understanding and further advancement in this domain\footnote{See Appendix~\ref{append:metrics} for all detailed metric description.}.

Motivated by these challenges, this paper systematically investigates the research question: \textit{What are the essential factors in crafting effective long-context multi-hop instruction datasets?}
To address this, inspired by recent advancement on Agent~\citep{hu2023tree,10.1145/3690642}, we propose a Multi-agent Interactive Multi-hop Generation (\texttt{MIMG}) framework.
First, to ensure data quality, a Quality Verification Agent is introduced to evaluate the quality of long-context samples throughout the process.
Second, for multi-hop reasoning, a Single-hop Question Generation Agent will be followed by a Multi-hop Question Merging Agent for stepwise synthesis of multi-hop queries.
Finally, to ensure diversity, Multiple Question Sampling strategies are proposed to reduce redundancy and promote variety.
To comprehensively examine the factors in long-context multi-hop data creation, we conduct extensive experiments, applying 17 strategies across 10 domains and 5 LLMs. As shown in Figure~\ref{fig:intro} (b), our method significantly improves data quality, yielding over 85\% multi-hop, high-quality, and non-duplicative samples. Notably, LLMs trained on the synthetic high-quality data show an average improvement of 7.54\%, even surpassing those LLMs trained on larger human-annotated datasets.

Overall, the main contributions are as follows:
\begin{itemize}[left=2pt,topsep=2pt,itemsep=1pt, parsep=2pt]
    \item
	We systematically explore strategies for generating high-quality multi-hop instruction data to identify unexplored but critical factors, that influence the quality of long-context data. These factors include scoring verifiers, question-then-answer generation, question-based sampling, and question-answer merging strategies.
	\item
	We introduce the Multi-Agent Interactive Multi-hop Generation (\texttt{MIMG}) framework, which enhances the quality and relevance of synthesized data through multiple agent interactions.
    \item 
	Our synthetic dataset, \texttt{LongMIT}, has shown superior performance across various long-context datasets. It not only improves long-context utilization but also surpasses larger human-labeled datasets, demonstrating its practical impact on advancing long-context LLMs.
\end{itemize}

\section{Framework}
Our framework consists of 4 main components: quality verification agent (QVA; $\S$~\ref{sec:quality-verification}), single-hop question generation agent (SQGA; $\S$~\ref{sec:question-generation}), multiple question sampling (MQS; $\S$~\ref{sec:question-sampling}), and multi-hop question merging agent (MQMA; $\S$~\ref{sec:question-merge}).
Specifically, the QVA is first designed as a validator to control and supervise the data quality at each stage. The SQGA then generates simple and direct single-hop questions. Next, MQS strategies expand on this by sampling questions that cover various documents, enhancing multi-hop instruction generation. Finally, the MQMA integrates these single-hop questions into the multi-hop questions, requiring multiple document information. The detailed architecture is illustrated in Figure~\ref{fig:framework}.

\subsection{Quality Verification Agent}
\label{sec:quality-verification}
The first module is the Quality Verification Agent, which globally supervises and ensures the quality of generated samples. This component involves two main processes:

\paragraph{Verification Strategy:}
This includes additional heuristic strategies to judge which samples should be contained as high-quality data.
Specifically, we explore two widely-used verification strategies:
\begin{itemize}[left=2pt,topsep=1pt,itemsep=1pt, parsep=5pt]
    \item \textbf{Scoring:} We prompt LLMs to assign continuous scores and determine a threshold using the validation set to filter high-quality data. Formally, given a sample $s$, the selection criterion is:
    \begin{equation}
        \!\!\mathcal{V}(s| \mathcal{M})\! =\! \begin{cases} 
            \texttt{Approved} & \mathcal{F}_S(s| \mathcal{M})\! > \!\theta; \\
            \texttt{Rejected} & \mathcal{F}_S(s| \mathcal{M})\! \leq \!\theta,
        \end{cases}
    \end{equation}
    where $\mathcal{F}_S(s| \mathcal{M})$ represents the score of sample $s$ based on model $ \mathcal{M}$, and $\theta$ is the threshold.
    \item \textbf{Classification:} We prompt LLMs to perform binary classification and retain only samples classified as high-quality. Formally, given a sample $s$, the selection criterion is:
    \begin{equation}
        \!\!\mathcal{V}(s| \mathcal{M})\! =\! \begin{cases} 
            \texttt{Approved} & \mathcal{F}_C(s| \mathcal{M})\! =\! 1; \\
            \texttt{Rejected} & \mathcal{F}_C(s| \mathcal{M})\! =\! 0,
        \end{cases}
    \end{equation}
    where $\mathcal{F}_C(s| \mathcal{M})$ represents the binary classification process of sample $s$.
\end{itemize}

\paragraph{Verification Condition:} 
This involves setting specific conditions $\mathcal{C}$ that both questions and answers must meet to be considered high-quality verification ($\mathcal{V}(s| \mathcal{M}, \mathcal{C})$). The process includes:
\begin{itemize}[left=2pt,topsep=1pt,itemsep=1pt, parsep=5pt]
\item \textbf{Criteria Perspectives:} Criteria include relevance to the document, clarity, factual accuracy, logical coherence, and complexity of the question and answer. Formally, these perspectives can be formulated as:
\begin{equation}
    \mathcal{C} = \{c_1, \dots, c_n\},
\end{equation}
where $c_i$ denotes the $i$-th criteria instruction. $n$ denotes the number of criteria perspectives.
\item \textbf{Auxiliary Context Information:} We integrate additional contextual instructions to enhance the model's accuracy and robustness, like guidelines. These conditions are formally represented as: 
\begin{equation}
    \mathcal{C} = \{c_1, \dots, c_n\} \oplus \texttt{Context},
\end{equation}
where the $\texttt{Context}$ denotes the context including auxiliary guidelines.
\item \textbf{Auxiliary Generation Information:} We enable the model to provide reasoning rationale during output generation and observe its effectiveness.
\begin{equation}
    \mathcal{C} = \{c_1, \dots, c_n\} \oplus I_R,
\end{equation}
where the $I_R$ denotes the instruction that can prompt LLM to generate rationales.
\end{itemize}

\subsection{Single-hop Question Generation Agent}
\label{sec:question-generation}
This phase generates single-hop questions and answers from individual documents through the following components:
\paragraph{Generation Backbone:}
We need a robust LLM to generate multiple single-hop questions and answers per document, ensuring diversity for data synthesis. Therefore, we evaluate various LLMs, both open- and closed-source, across different scales.
\paragraph{Generation Strategy:}
Our strategy employs structured techniques to extract potential questions:

\begin{itemize}[left=2pt,topsep=1pt,itemsep=1pt, parsep=5pt]
\item \textbf{Rationale-based Question Generation:} Chain-of-Thought (CoT) can effectively enhances performance on long-context tasks~\citep{li-etal-2024-making}. Hence, we explore whether generating rationale-supported questions improves the model's grasp of a document's reasoning structure.
\item \textbf{Question-Answering Generation Order:} Furthermore, we assess whether order affects output quality. Generating questions before answers may simplify reasoning and enhance output clarity compared to simultaneous generation.
\end{itemize}

\subsection{Multiple Question Sampling}
\label{sec:question-sampling}
To further optimize the diversity of generated samples, we introduce
MQS strategy, which constructs multi-hop questions by sampling and combining questions from various documents. This approach consists of two key strategies:

\paragraph{Retrieval Strategy:}
This strategy selects relevant questions and documents for multi-hop question generation. By leveraging relevance sampling, it constructs a question-semantic relevance matrix to assess semantic connections across different documents, guiding the sampling process. The strategy comprises:

\begin{itemize}[left=2pt,topsep=1pt,itemsep=1pt, parsep=5pt]
    \item \textbf{Probability-Based Sampling:} This method samples data based on probability-based document relevance, which is calculated as the frequency of keywords related to the questions, like BM25~\citep{robertson1995okapi,robertson2009probabilistic}, and LDA~\citep{hoffman2010online}.
    \item \textbf{Semantic-Based Sampling:} This approach assesses the relevance by analyzing the semantic similarity between questions and documents, like embedding similarity.
\end{itemize}

\paragraph{Sampling Strategy:}
Based on the relevance matrix, the most related questions should be selected to form a coherent, contextually rich multi-hop question.  The strategy includes:

\begin{itemize}[left=2pt,topsep=1pt,itemsep=1pt, parsep=5pt]
    \item \textbf{Intra-Document Sampling:} It focuses on selecting questions within the same document to ensure internal coherent multi-hop data.
    \item \textbf{Inter-Document Sampling:} This strategy involves selecting questions from different documents to ensure a broader contextual coverage.
\end{itemize}

\subsection{Multi-hop Question Merging Agent}
\label{sec:question-merge}
The final step merges sampled questions into coherent multi-hop questions, involving two modules:

\paragraph{Merging Backbone:}
We should employ LLMs to synthesize sampled questions and answers into meaningful multi-hop queries. To investigate this, we conduct the following exploration:

\paragraph{Merging Strategy:} 
This strategy applies rules and heuristics to maintain logical and contextual consistency. It includes:
\begin{itemize}[left=2pt,topsep=1pt,itemsep=1pt, parsep=5pt]
    \item \textbf{Document-Based Merging:} To further reduce input tokens, we explore whether incorporating long documents enhances merging performance. Formally, the process is:
    \begin{equation}
        {Q_m} = \mathcal{M}(Q_1, Q_2, \ldots, Q_n|C),
    \end{equation}
    where $Q_1, Q_2, \ldots, Q_n$ are sampled single-hop questions, $Q_m$ is the merged multi-hop query, and $C$ denotes whether documents are used.
    \item \textbf{Rationale-Based Merging:} To preserve the meaning and context of the individual questions, this method leverages the reasoning rationale behind the original questions to guide their merging process, which can be expressed as:
    \begin{equation}
        R \oplus {Q_m} = \mathcal{M}(Q_1, Q_2, \ldots, Q_n),
    \end{equation}
    where $R$ represents the underlying rationale, and $\oplus$ denotes the connecting elements in the generated response.
\end{itemize}

Additionally, we explore intra-document and inter-document multi-hop instruction samples for diverse scenarios.

\begin{figure}[t]
	\centering
	\includegraphics[width=0.48\textwidth]{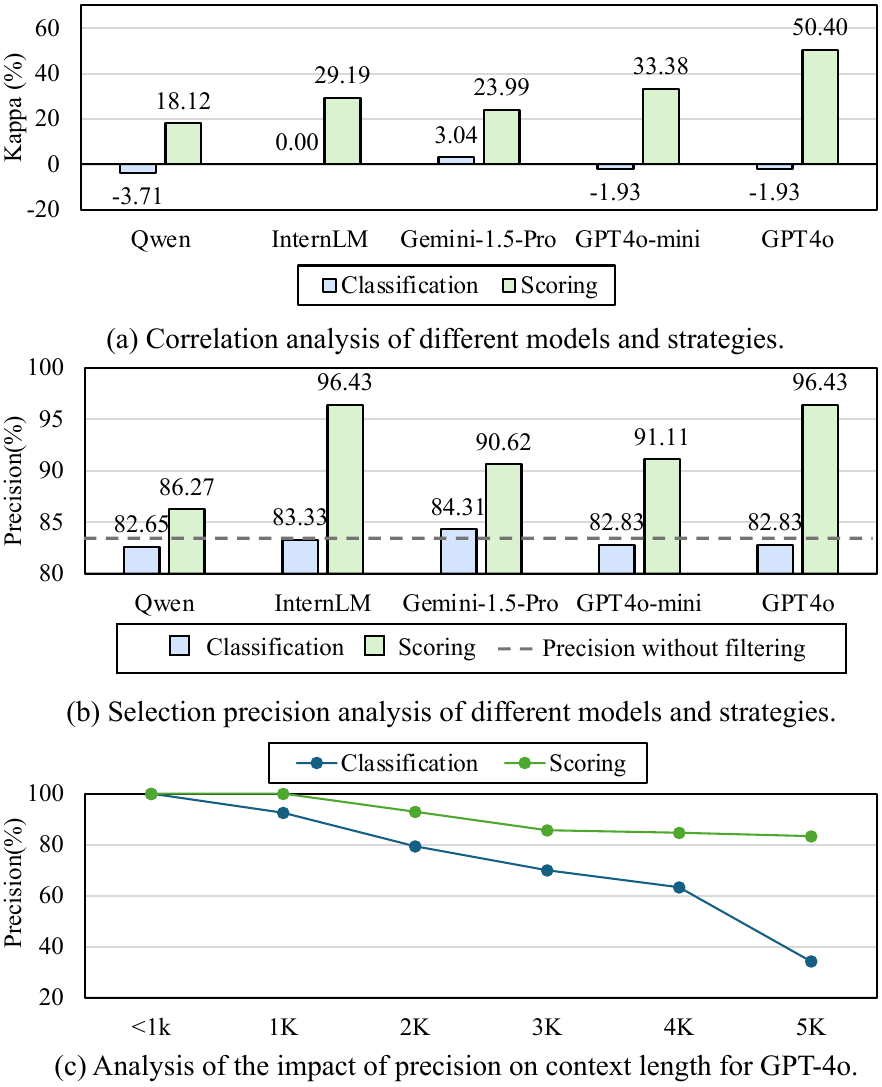}
	\caption{The analysis of different verification strategies in quality verification, where includes 5 models: Qwen2-72B-Instruct~\citep{yang2024qwen2}; InternLM2-20B~\citep{cai2024internlm2}; Gemini-1.5-Pro~\citep{reid2024gemini}; GPT-4o-mini and GPT-4o~\citep{achiam2023gpt}.
	}
	\label{fig:verify-strategy}
\end{figure}
\begin{figure*}[t]
	\centering
	\includegraphics[width=0.98\textwidth]{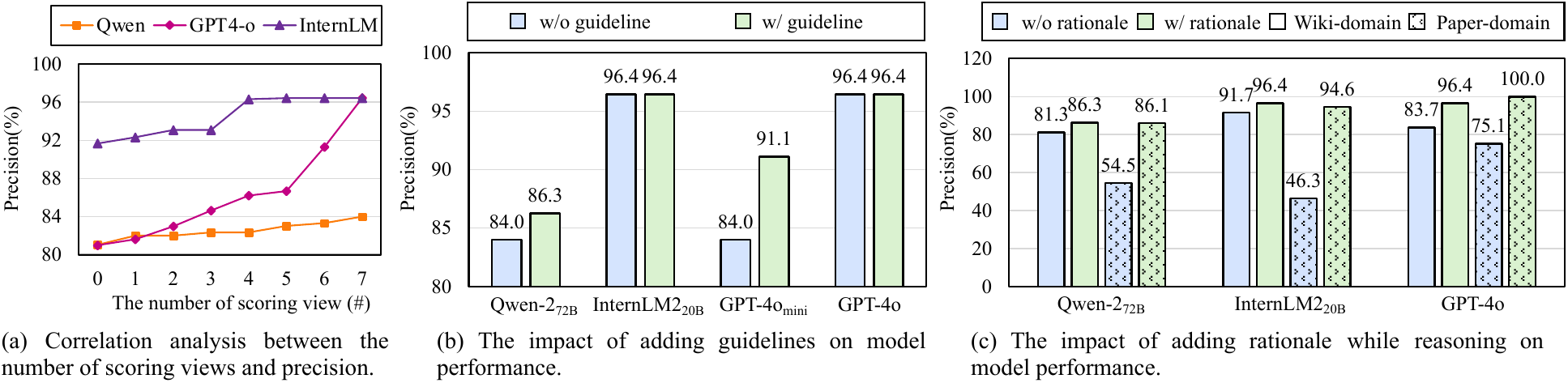}
	\caption{The analysis of different verification conditions on quality verification.
	}
	\label{fig:verify-condition}
\end{figure*}
\section{Exploration}
This section examines the framework components aimed at enhancing data quality, including verification strategies in QVA (\S \ref{sec:qva-exp}), generation strategies in SQGA (\S \ref{sec:sqga-exp}), sampling approaches in MQS (\S \ref{sec:mqs-exp}), and merging strategies in MQMA (\S \ref{sec:mhqma-exp}).
\subsection{Quality Verification Agent}
\label{sec:qva-exp}
\subsubsection{Verification Strategy}
\label{sec:qva-v-exp}
Currently, the most commonly used model verification strategies are scoring and direct classification. We assessed the consistency and accuracy of both methods by comparing them with human annotations in a sample analysis of long-context data.

\paragraph{Scoring is a better verification strategy compared with classification.} As shown in Figure~\ref{fig:verify-strategy} (a), the scoring method significantly outperforms binary classification, yielding higher kappa and precision scores. This statistical improvement indicates that scoring better captures the subtleties of human judgment. These findings align with research in short-context scenarios~\citep{fu-etal-2024-gptscore}, highlighting the broader applicability of scoring methods across different context lengths.

\paragraph{LLM is not a long-context annotator but a good selector.} As depicted in Figure~\ref{fig:verify-strategy} (a),
unlike their performance in short-context verification~\citep{wang-etal-2023-chatgpt,fu-etal-2024-gptscore}, LLMs show minimal agreement with human annotators in long-context situations, reflected in low kappa scores. This suggests challenges in maintaining consistent annotations due to cognitive load and interpretive variations over extended data. However, as Figure~\ref{fig:verify-strategy} (b) shows, LLMs maintain near-perfect precision, demonstrating their strong ability to identify and select relevant information. This highlights LLMs' potential as effective tools for data filtering and prioritization in long-context environments, in contrast to their role as accurate annotators in short-context settings.

\paragraph{Scoring alleviates the long context bias but classification does not.} We further examine why classification performs poorly in long-context scenarios by analyzing precision across different context lengths. Figure~\ref{fig:verify-strategy} (c) illustrates that scoring consistently achieves higher precision and robustness in longer contexts, explaining the suboptimal performance of classification in these cases. Based on these findings, subsequent experiments will adopt the scoring strategy, using verifier precision and the data retention ratio to assess data quality. More discussion can be seen in Appendix~\ref{append:score-class}.

\subsubsection{Verification Conditions}
To deeply understand what factors affect the verification of long text data quality, we further explored from three perspectives: scoring perspective, guidelines, and whether rationale is included for scoring.

\paragraph{More scoring criteria reduce long-context bias.} As shown in Figure~\ref{fig:verify-condition} (a), incorporating more scoring criteria enhances the accuracy and robustness of filtering long-context data. Unlike short contexts, long contexts are prone to judgment bias. When fewer than three criteria are used, performance improvements are limited, with models often overestimating irrelevant samples. Increasing the number of criteria improves labeling accuracy, reducing biases linked to longer contexts (see Appendix~\ref{append:qva} for further details).

\paragraph{Effective verifiers adhere to annotation standards aligned with human judgment without guidelines.} To assess whether incorporating additional guidelines improves verification performance, we analyze the effectiveness of the method of integrating guidelines. Figure~\ref{fig:verify-condition} (b) reveals that advanced verifiers do not require supplementary guideline information during annotation. This suggests that effective verifiers naturally adhere to annotation standards that align with human judgment.

\paragraph{Incorporating rationale enhances robustness in diverse long contexts.} By examining CoT~\citep{NEURIPS2022_9d560961,qin-etal-2023-cross}, in various domains like wiki knowledge and paper analysis, we observe that incorporating rationale improves model performance across diverse long-context scenarios. As shown in Figure~\ref{fig:verify-condition} (c), without rationale, performance drops by more than 8.6\% across different domains. However, adding rationale during validation minimizes performance variation, with precision fluctuations limited to 1.8\%.

\subsection{Single-hop Question Generation Agent} 
\label{sec:sqga-exp}
\subsubsection{Generation Backbone}

In practice, effective LLMs must be able to synthesize high-quality data. Thus, we evaluated several widely used LLMs for single-hop data synthesis.

\paragraph{Open-source LLMs effectively generate single-hop questions.} As shown in Figure~\ref{fig:single-hop} (a), smaller open-source LLMs exhibit high retention rates and cost-effectiveness, demonstrating their ability to generate single-hop questions from a given context.

\paragraph{Stronger LLMs can generate better single-hop question generation but higher cost.} As shown in Figure~\ref{fig:single-hop} (a), more advanced LLMs enhance data retention and question quality. However, these improvements are not cost-proportional, raising concerns about their economic feasibility for single-hop question generation.

\subsubsection{Generation Strategy} 
\begin{figure}[t]
	\centering
	\includegraphics[width=0.48\textwidth]{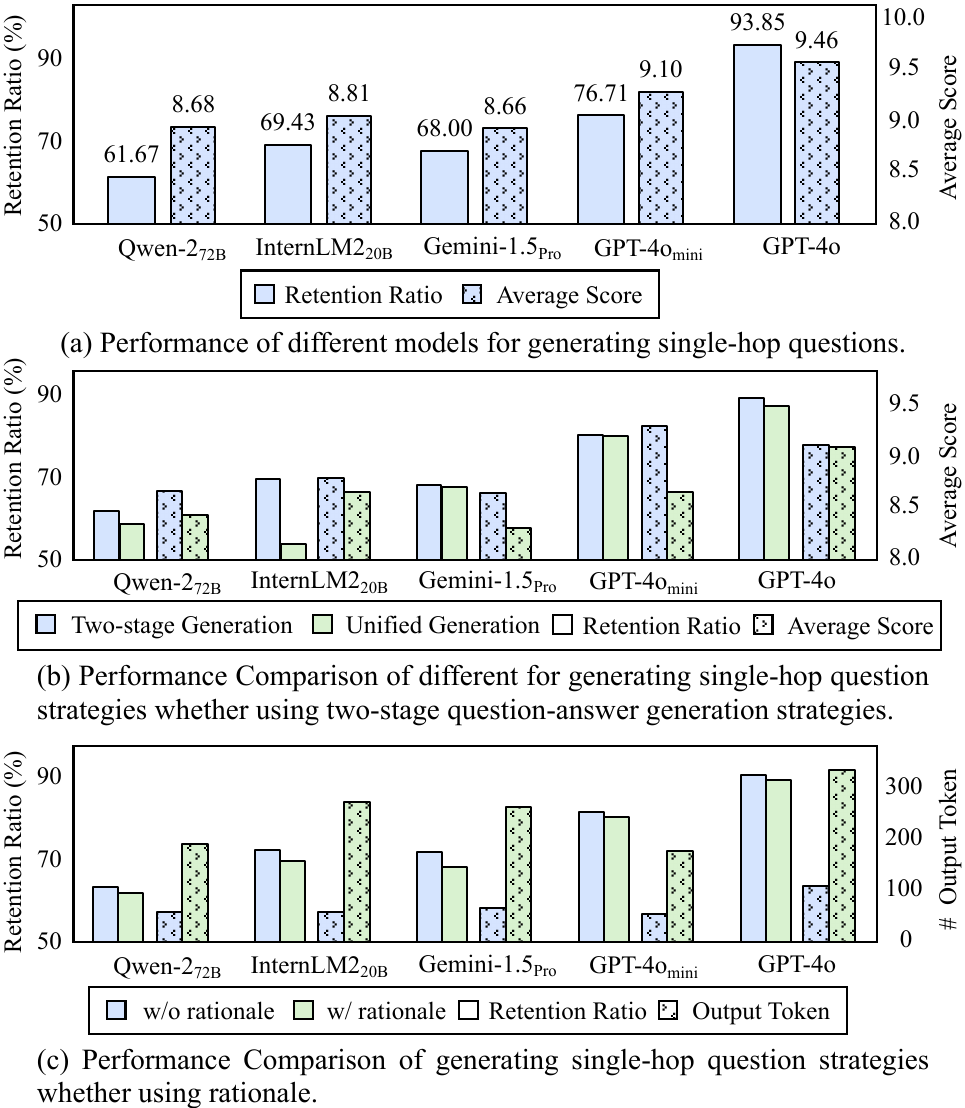}
	\caption{The analysis of generation backbone and generation strategies in SQGA.
	}
	\label{fig:single-hop}
\end{figure}

Furthermore, we explore whether a question-then-answer approach, supported by rationale, improves the quality of synthetic single-hop questions.
\paragraph{Question-then-answering works better than generating data from scratch.}
To evaluate the effectiveness of single versus multi-stage generation, we compare two strategies: unified question-answer and question-then-answer. As shown in Figure~\ref{fig:single-hop} (b), generating the question first significantly improves data quality, particularly for open-sourced LLMs, increasing retention and quality scores. For implementation details, refer to Appendix~\ref{append:sqga}.

\begin{figure}[t]
	\centering
	\includegraphics[width=0.45\textwidth]{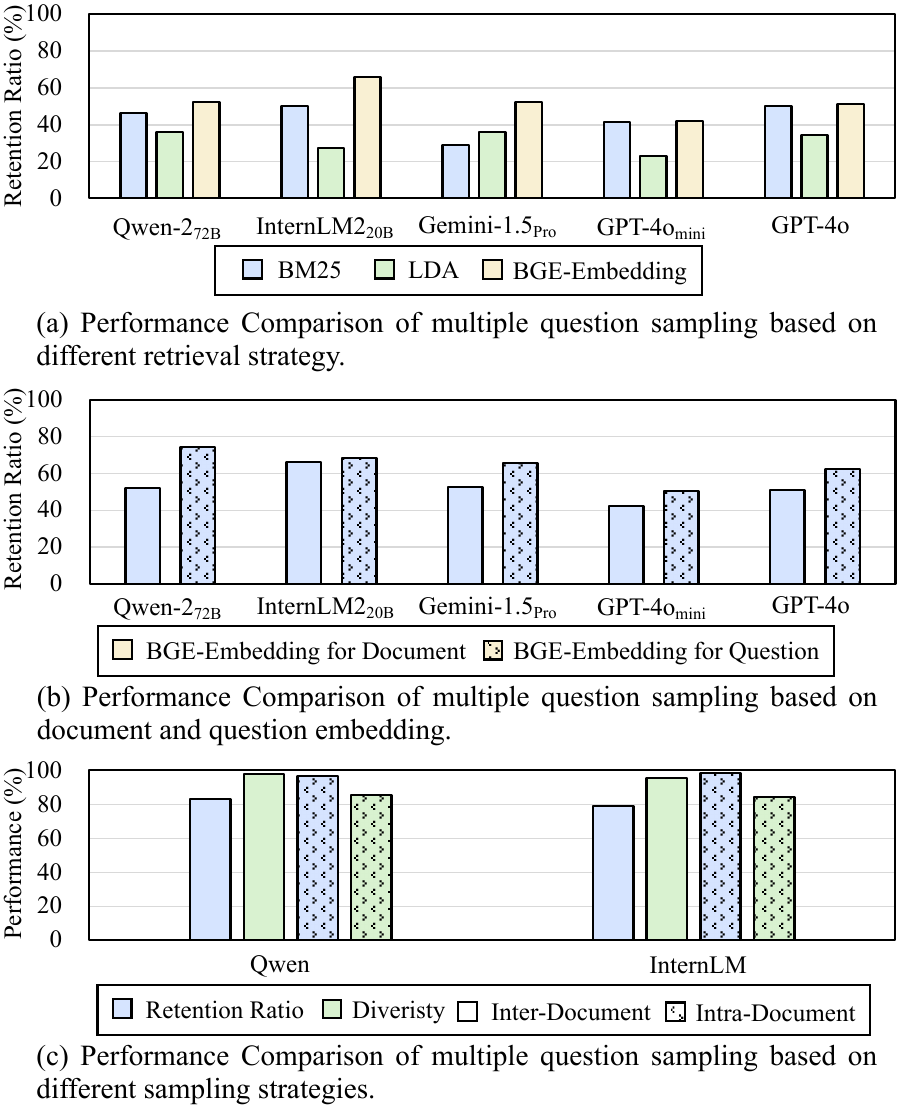}
	\caption{The analysis of multiple question sampling.
	}
	\label{fig:multi-sampling}
\end{figure}

\paragraph{Generating with rationale can improve the generated quality but much higher token cost.} As illustrated in Figure~\ref{fig:single-hop} (c), adding rationale makes questions more relevant and insightful with higher quality. However, the improvement brought by the rationale is minimal, while the token consumption triples, making it economically inefficient.

\subsection{Multiple Question Sampling}
\label{sec:mqs-exp}
\subsubsection{Retrieval Strategy}
This strategy focuses on identifying relevant documents and constructing a semantic relevance matrix to guide sampling. Observations include:
\paragraph{Embedding similarity is critical for multi-question sampling.} We evaluate the effectiveness of various similarity measures using three metrics: embedding similarity (with BGE embeddings~\citep{bge_embedding}), BM25, and LDA. As shown in Figure~\ref{fig:multi-sampling} (a), BGE embeddings enable the model to select more relevant questions, thereby improving sample quality.

\paragraph{Question similarity outweighs document similarity.} We further investigate the factors affecting sample quality. Figure~\ref{fig:multi-sampling} (b) reveals that question-based sampling significantly outperforms document-based approaches, as questions provide greater contextual relevance.

\begin{figure}[t]
	\centering
	\includegraphics[width=0.48\textwidth]{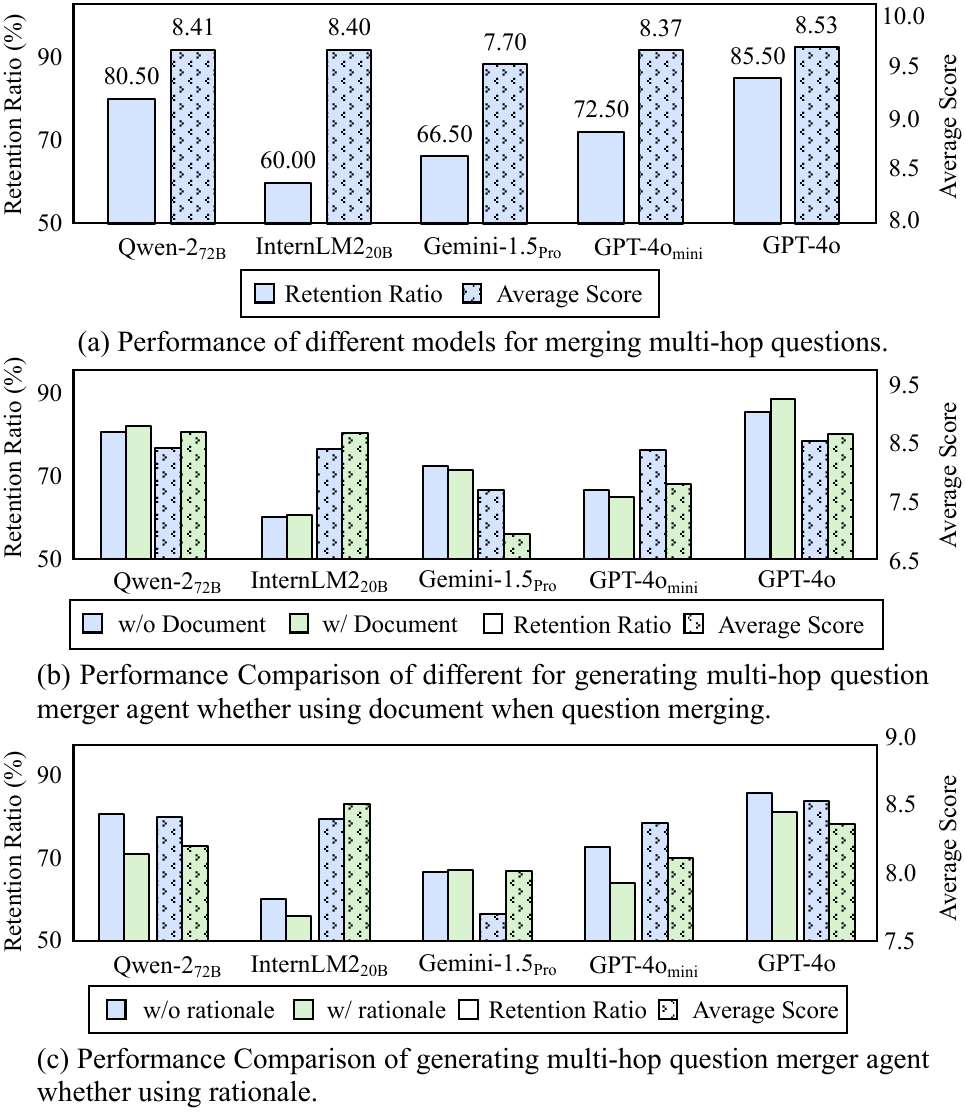}
	\caption{The analysis of multi-hop question merging agent.
	}
	\label{fig:multi-hop}
\end{figure}

\subsubsection{Sampling Strategy}

This approach selects semantically related and complementary questions from within and across documents to create coherent, contextually rich multi-hop questions.

\paragraph{Intra-Document generates better quality but less diversity.} As shown in Figure~\ref{fig:multi-sampling} (c), sampling questions within a single document yields more coherent and contextually aligned questions. However, this method may reduce diversity, as the questions are all drawn from the same source.
\paragraph{Inter-Document generates less quality but more diversity.} As demonstrated in Figure~\ref{fig:multi-sampling} (c), sampling from multiple documents introduces a wider range of topics, enhancing diversity. However, this broader scope may diminish coherence and relevance due to the larger topic gaps.

\subsection{Multi-hop Question Merging Agent}
\label{sec:mhqma-exp}
\subsubsection{Merging Backbone} We use LLMs to merge sampled questions and answers into coherent multi-hop versions, ensuring logical consistency and contextual accuracy with the aid of 5 classic LLMs. The key observations are as follows:
\paragraph{Open-sourced LLMs can well merge multi-hop question generation.} As shown in Figure~\ref{fig:multi-hop} (a), all models demonstrate strong capabilities in handling complex question-generation tasks requiring multi-step reasoning or information integration.

\subsubsection{Merging Strategy}
\paragraph{Question-answer pairs are enough for multi-hop instruction merging compared with documents.} To minimize input tokens, we examine whether long documents are necessary for improving merging performance. Figure~\ref{fig:multi-hop} (b) shows that adding documents does not consistently enhance performance and instead increases input tokens. Thus, simple question-answer pairs are sufficient for effective multi-hop merging.

\paragraph{Merging with rationale can not improve the merging quality.} Although generating content with rationales generally enhances quality~\citep{qin-etal-2023-cross,qin2024large}, Figure~\ref{fig:multi-hop} (c) demonstrates that, unlike single-hop generation, rationales in multi-hop generation fail to contribute to coherent and logical question formation. Our analysis further reveals that large models often misinterpret rationales in queries and merging strategies, leading to frequent failures in CoT reasoning. Therefore, multi-hop synthesis should avoid using rationales.
\begin{table*}[t]
	\centering
    \begin{adjustbox}{width=0.98\textwidth}
        \begin{tabular}{lccccccccc}
            \toprule
            Model & NarrativeQA & 2WikiMQA & DuReader & HotpotQA & MultifieldQA$_{en}$ & MultifieldQA$_{zh}$ & MuSiQue & Qasper & AVG \\
            \midrule
            \rowcolor{gray!8}\multicolumn{10}{c}{InternLM2-1.8B~\citep{cai2024internlm2}}\\
            \midrule
            \ \ \texttt{+ChatQA2} & 18.50 & 35.00 & 29.00 & 46.00 & 64.00 & 58.00 & 19.50 & 38.50 & 38.56 \\
            \ \ \texttt{+LongAlign} & 25.00 & 33.00 & 25.00 & 49.50 & \textbf{76.00} & 67.50 & 24.50 & 44.00 & 43.06 \\
            \ \ \texttt{+LongAlpaca} & 25.00 & 23.50 & 29.00 & 49.50 & 70.00 & 67.00 & 24.50 & 45.00 & 41.69 \\
            \ \ \texttt{+NQ} & 17.00 & 25.50 & 33.50 & 35.00 & 60.00 & 67.00 & 14.50 & 44.00 & 37.06 \\
            \ \ \texttt{+LongMIT} & \textbf{26.00} & \textbf{35.50} & \textbf{60.00} & \textbf{56.00} & 75.33 & \textbf{75.50} & \textbf{29.00} & \textbf{47.50} & \textbf{50.60}   \\
            \midrule
            \rowcolor{gray!8}\multicolumn{10}{c}{LLaMA3-8B~\citep{dubey2024llama}}\\
            \midrule
            \ \ \texttt{+ChatQA2} & 24.00 & 41.00 & 50.00 & 49.00 & 64.00 & 69.00 & 26.00 & 51.50 & 46.81  \\
            \ \ \texttt{+LongAlign} & 29.00 & 44.50 & 56.50 & 56.50 & 79.33 & 80.50 & 21.50 & 55.50 & 52.92 \\
            \ \ \texttt{+LongAlpaca} & 18.00 & 50.00 & 48.00 & 55.50 & 76.67 & 80.00 & 27.50 & \textbf{60.50} & 52.02  \\
            \ \ \texttt{+NQ} & 21.00 & 42.00 & 63.00 & 59.50 & 78.00 & 74.00 & 29.00 & 54.00 & 52.56  \\
            \ \ \texttt{+LongMIT} & \textbf{36.50} & \textbf{67.50} & \textbf{74.00} & \textbf{71.00} & \textbf{87.33} & \textbf{84.50} & \textbf{39.50} & 54.00 & \textbf{64.29}   \\
            \midrule
            \rowcolor{gray!8}\multicolumn{10}{c}{InternLM2-7B~\citep{cai2024internlm2}}\\
            \midrule
            \ \ \texttt{+ChatQA2} & 31.00 & 42.00 & 38.50 & 61.00 & 70.67 & 33.00 & 28.50 & 53.00 & 44.71  \\
            \ \ \texttt{+LongAlign} &45.00 & 40.00 & 60.00 & 65.50 & 74.67 & 86.00 & 34.00 & 56.50 & 57.71  \\
            \ \ \texttt{+LongAlpaca} & 45.00 & 50.50 & 44.00 & 64.50 & 75.33 & 47.50 & 35.50 & 56.50 & 52.35  \\
            \ \ \texttt{+NQ} & 12.50 & 37.50 & 61.50 & 45.50 & 75.33 & 77.00 & 21.00 & 57.50 & 48.47 \\
            \ \ \texttt{+LongMIT} & \textbf{46.50} & \textbf{57.00} & \textbf{74.00} & \textbf{73.00} & \textbf{91.33} & \textbf{91.00} & \textbf{45.00} & \textbf{62.00} & \textbf{67.48}  \\
            \bottomrule
        \end{tabular}
    \end{adjustbox}
    \caption{Main accuracy results by evaluation by GPT-4o, where all benchmarks come from the LongBench~\citep{bai2024longbench}. More evaluation results on Ruler~\citep{hsieh2024ruler} are shown in Table~\ref{exp:ruler-exp}.}
    \label{exp:main-exp}
\end{table*}
\section{Data Utilization}

\subsection{Data Synthesis Efficiency}
\begin{figure}[t]
    \centering
	\includegraphics[width=0.48\textwidth]{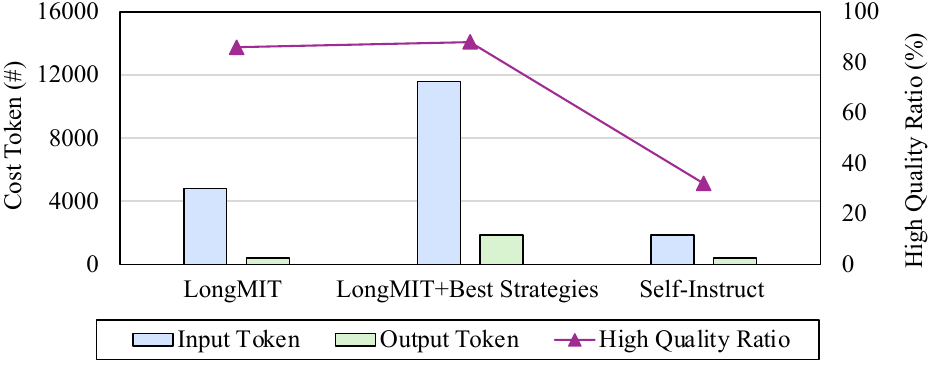}
	\caption{Comparison of the quality and token consumption on different generation strategies.
	}
	\label{fig:overall-cost}
\end{figure}
Given the high cost of data generation, we balance cost and data quality in synthesizing long multi-hop instruction tuning (\texttt{LongMIT}) datasets (See Appendix~\ref{append:data-construct} for more data details). To assess this balance, we compare the proportion of high-quality data and token cost for 200 samples generated under different strategies.
At the input level, our interactions increase the token count by approximately 3k tokens (roughly 2.5 times improvement), which is negligible in comparison to the substantial volume of long-context documents provided, averaging 70k tokens as shown in Figure 9.
At the output level, the increase in token consumption is almost no noticeable increase, with an average increase of fewer than 0.5k tokens.
Despite this minimal change in token usage, the quality of our results has improved nearly fourfold.
As shown in Figure~\ref{fig:overall-cost}, strategies using open-source LLMs achieve a high-quality proportion comparable to the best strategies, at only one-third of the token cost. Additionally, our approach improves data quality with minimal token expense compared to traditional methods. Then, we will pad the context with additional documents to the target length to create sufficiently long samples. For further experimental results and details, see Appendix~\ref{append:best-precision}. Moreover, additional exploration for long code instruction data generation can be found in Appendix~\ref{append:code}.

\subsection{The Results of Instruction-Tuning}
We conduct instruction-tuning on the synthesized data to evaluate its utility. As shown in Table~\ref{exp:main-exp}, our data significantly improves long-context QA capabilities across various LLMs, with an average gain of at least 7.54\%. Multi-hop benchmarks such as 2WikiMQA~\citep{ho-etal-2020-constructing}, MuSiQue~\citep{10.1162/tacl_a_00475}, and HotpotQA~\citep{yang-etal-2018-hotpotqa} show more notable improvements.  Detailed training procedures are in Appendix~\ref{append:train}. Furthermore, in Appendix~\ref{append:case}, the high quality and logical complexity of this data enable the model to generalize to single-hop tasks not encountered during instruction tuning, confirming the reliability of \texttt{MIMG}.

\subsection{Scaling Analysis}
\subsubsection{Data Scaling Analysis} To evaluate how the size of high-quality data affects model performance, we experiment on LLaMA3-8B~\citep{dubey2024llama} by varying the training data volume. The results, depicted in Figure~\ref{fig:data-size-affect} in Appendix, show a clear relationship between the size of data and the performance. As the dataset size increases, model performance adjusts accordingly, demonstrating the significance of high-quality data scaling in enhancing the model efficacy.

\subsubsection{Hop Scaling Analysis}
To assess the impact of multi-hop data on model performance, we increased the number of hops in the dataset while keeping the training data volume constant. This approach isolated the effect of multi-hop reasoning on model outcomes. As indicated in Figure~\ref{fig:hop-affect} in Appendix, there is a clear positive correlation between the number of hops and model performance. The data demonstrate that with more hops, the model achieves higher accuracy and robustness, demonstrating the effectiveness of using high-quality multi-hop data to enhance the model's capability for complex reasoning tasks.
\section{Related Work}
Recent efforts have aimed to enhance the performance of LLMs in handling longer contexts~\citep{hu2024hiagent,liu2025survey}. LongLLaMA~\citep{xiong2023effective} demonstrates the impact of incorporating long text data during various pre-training stages. Then, LLaMA2-80K~\citep{fu2024data} highlights the significance of using a domain-balanced, upsampled long text corpus to improve long text capabilities, requiring only a 5B-token corpus for effective comprehension. Furthermore, ICLM~\citep{shi2024incontext} enhances long-text reasoning by transforming pre-training data into knowledge graphs and splicing adjacent documents.
To improve the model's ability to follow long text instructions, LongAlpaca~\citep{chen2024longlora} combines a 9K paper question-answering (QA) corpus with 3K short instruction samples. In contrast, LongAlign~\citep{bai2024long} utilizes Claude~\citep{anthropic2023claude} to produce 10K QA pairs for training. Additionally, ChatQA~\citep{liu2024chatqa} enhances long-context QA performance by incorporating manually annotated data. Building on these approaches, ChatQA2~\citep{xu2024chatqa} further incorporate existing long-text datasets, such as Natural Questions (NQ)~\citep{kwiatkowski2019natural}.

The method closest dataset is Quest~\citep{gao2024quest}, which constructs QA pairs from spliced documents, resulting in a close-sourced single-hop QA corpus. In contrast, our approach models document correlations first, then create multi-hop QA pairs using related intra-document data. Additionally, we offer systematic analysis, open-source datasets, and significantly improved models.
\section{Conclusion}
In conclusion, our proposed Multi-agent Interactive Multi-hop Generation (\texttt{MIMG}) framework, including a quality verification agent, a single-hop question generation agent, a multiple question sampling strategy, and a multi-hop question merger agent, achieves high-quality, diverse instruction data. Our experiments show that this synthetic data notably enhances performance, even surpassing larger human-annotated data, highlighting the effectiveness of our approaches.

\section*{Limitations}
Due to the high costs associated with large-scale distillation training experiments on GPT-4-based methods, we focus our evaluation efforts on a small-scale assessment conducted through manual evaluation. To ensure robustness, the manual annotation is performed on a sample of 200 items. While the sample size is manageable, we acknowledge the potential for minor unavoidable biases inherent in random sampling. Moreover, considering the  nature of randomization for LLMs, it is quite hard for us to strictly control LLMs to generate the same question for quality comparison.

\vspace{-2mm}\section*{Ethical Considerations}\vspace{-1mm}
Participants are recruited from universities across China, and all must have passed the CET-6 exam or achieved an IELTS score of 6 or higher. While participants come from diverse regions, we minimize the impact of national biases by focusing primarily on long context data. All annotators provided informed consent and were compensated above the local minimum wage. Additionally, no IRB review was required for the study.

The annotation process starts with an onboarding test, where participants answer 20 example questions. They receive \$42 for this phase, designed to familiarize them with the task. Annotators are then paid \$5 per hour, with a total of approximately 120 human hours dedicated to manual annotations. Overall, three experts are involved in the annotation and verification stages.
\vspace{-2mm}\section*{Acknowledgments}\vspace{-1mm}
Thanks to Hanqi Li, Yifei Yang and Da Ma for their constructive feedbacks.

\bibliography{custom}

\appendix
\newpage
\section*{Appendix}

\section{Metrics Utilized in Exploration}
\label{append:metrics}
\subsection{Metric Definition \& Implementation}
\subsubsection{Diversity}
Diversity measures the frequency of samples with the same semantics appearing in the data. In annotation, annotators sequentially identify whether new samples are semantically equivalent to previously annotated ones.

High diversity indicates a broad range of samples, ensuring that annotations do not repeat similar or identical meanings. This is essential for creating a dataset that represents various use cases and scenarios, covering a wide array of semantic topics. A diverse dataset contributes to a more robust LLM by capturing the nuances of language, context, and conceptual meaning across different long-context samples.

\subsubsection{Multi-Hop}
Multi-hop refers to the need for multiple information connections and integrations from various sources when handling complex samples. Here, annotation tasks require annotators to assess a query's needs by utilizing information from several contextual documents.

Effective multi-hop datasets pose questions that cannot be answered with a single data point but demand the combination of multiple facts or steps to reach the correct answer. Such reasoning is vital in real-world long-context tasks, such as answering questions that require complex deductions or understanding interconnected pieces of information.

\subsubsection{High-Quality}
High-quality annotations denote the accuracy, consistency, and relevance of the synthesis data. In high-quality datasets, annotators are required to judge whether each sample is precise and reliable, minimizing errors or inconsistencies.

Annotators must ensure that the data they provide accurately reflects the meaning or intent of the task at hand. In long-context NLP, high-quality data is crucial for developing models that make accurate predictions, recognize subtle patterns, and perform effectively in real-world scenarios.

\begin{table}[t]
	\centering
    \begin{adjustbox}{width=0.48\textwidth}
        \begin{tabular}{l|ccc}
            \toprule
             
            Dataset Name & High-quality & Diversity & Multi-hop \\
            \midrule
            
            LongAlpaca~\citep{chen2024longlora} & 70.3 & 87.9 & 50.7 \\
            LongAlign~\citep{bai2024long} & 87.7 & 83.8 & 52.6 \\
            ChatQA2~\citep{xu2024chatqa} & 83.1 & 80.3 & 30.4 \\
            NQ~\citep{kwiatkowski2019natural} & 90.2 & 63.2 & 10.3 \\
            \texttt{LongMIT} (Ours) & 94.8 & 88.2 & 94.8 \\
            \bottomrule
        \end{tabular}
    \end{adjustbox}
    \caption{Results of the data quality of different instruction datasets, reporting the data quality score as the average of three independent manual evaluations.\vspace{-4mm}}
    \label{exp:quality-relation}
\end{table}

\subsection{The Impact of Different Metric}
Furthermore, we conduct sampling and annotation of several instruction-tuning datasets based on three metrics. A comparison with Table~\ref{exp:quality-relation} shows that the results require considering multiple factors and the specific needs of each task: NQ provides high-quality data, enhancing its effectiveness for in-domain tasks such as DuReader. However, its lack of diversity limits its performance in out-of-domain scenarios, which lowers its overall accuracy. In contrast, while LongAlign offers greater diversity, its quality is relatively lower, weakening its fundamental comprehension abilities. This is reflected in its significantly poorer performance on basic tasks like DuReader, resulting in suboptimal overall performance.

For the metric view, the conclusions are as follows: (1) High-quality annotations notably improve model performance in foundational long-context comprehension tasks, such as DuReader. (2) Annotation diversity is essential for enhancing model performance across a wide range of tasks. (3) The multi-hop property is particularly important for complex reasoning tasks that require integrating multiple long-context clues, as seen in datasets like HotpotQA and MusiQue.

In summary, we argue that quality ensures strong performance in fundamental long-context tasks, multi-hop promotes the complex multi-hop capabilities, while diversity improves performance across various fields by leveraging these basic capabilities. 

\subsection{Automatic Metrics}
\subsubsection{Quality Score}
Several studies have shown that large-scale models align with human judgment when generating quality scores~\citep{chen2023exploring, chang2024survey, lee2024unleashing}. These models also perform well in specialized domains, such as medicine, where they show strong consistency with expert evaluations~\citep{haim2024ai}. Motivated by these findings, we prompt LLMs to automatically generate quality scores for scalable and effective long-context data quality verification.

To analyze the effectiveness of quality scores in long-context scenarios, we analyze the consistency between quality scores and human assessments. Specifically, Figure~\ref{fig:verify-strategy} (a) demonstrates that the Kappa coefficient for the agreement between the scoring mechanism and human evaluators exceeds 0.50, significantly surpassing the performance of classification strategies that directly label items as high or low quality. Furthermore, Figure~\ref{fig:verify-strategy} (b) shows that the scoring mechanism achieves a precision level of 96.43\% relative to human evaluators, underscoring its effectiveness as a robust data screening strategy.

\subsubsection{Retention Rate}
Retention rate refers to the proportion of data retained after being filtered by a quality verification agent based on a specified threshold. This concept arises from our observation that LLMs excel more as selectors than as annotators in Sec.~\ref{sec:qva-v-exp}, demonstrating higher retention rates and better alignment with human judgment.
In complex or subtle situations, LLM judgments may vary, which can impact data interpretability and reliability. Therefore, we select models and strategies that closely align with human consistency, which achieves an almost perfect precision rate of 96.43\%.

For implementation, we use a threshold of 8.5, determined through an internally annotated verification set of 200 items. The threshold yielding the highest precision in this set was chosen as the standard parameter for future evaluations.

Regarding the ``average score'', our strategy assigns a quality score from 0 to 10 for each data point. After evaluating the entire dataset, we calculate the average score by averaging all individual quality scores.

\begin{figure*}[t]
	\centering
	\includegraphics[width=0.98\textwidth]{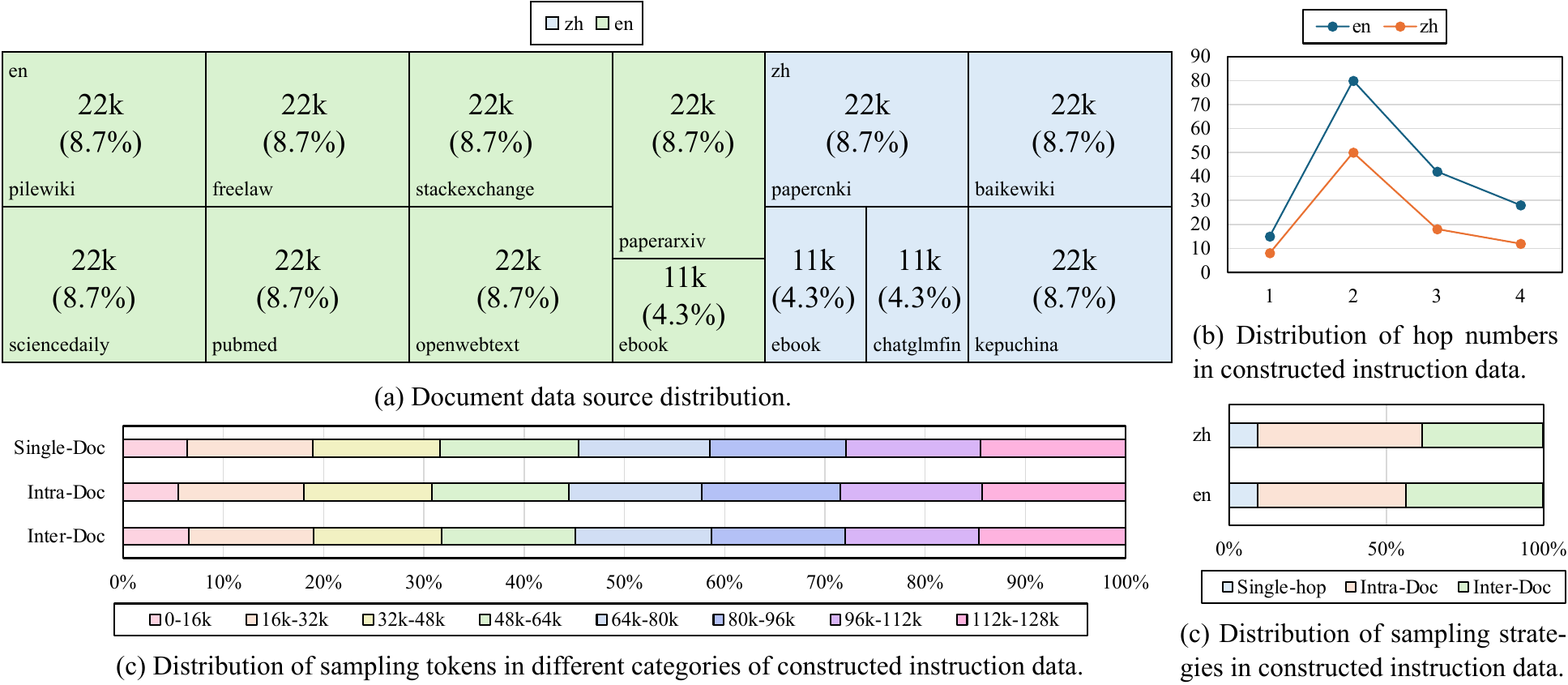}
	\caption{The analysis of constructed dataset distribution.
	}
	\label{fig:data-analysis}
\end{figure*}

\section{Discussion of Scoring versus Classification}
\label{append:score-class}
Although the document inputs are relatively long, averaging approximately 3K tokens, the classifier still assigns a ``high-quality'' label to nearly every sample of this length, even when we introduce multiple reference criteria. We believe that, once inputs exceed this length threshold, the classifier’s baseline score effectively rises above five, so any score over five is automatically deemed high quality. As a result, the classifier’s judgments lose discriminative power: every long sample is labeled ``high-quality''.

By contrast, the scoring approach provides a more nuanced ranking. Even if all samples are of generally high quality, a score of 0.9 is still recognized as slightly better than 0.8. This finer distinction more closely reflects human evaluation.

\section{Data Construction Details}
\label{append:data-construct}
\subsection{Dataset Construction Pipeline Discussion}
The current data in this field relies on Self-Instruct but lacks a systematic framework. Notably, many existing long-text works are essentially subsets of our \texttt{MIMG}. For instance, the prompts of Self-Instruction and those of the Single-hop Question Generation Agent are nearly identical. Additionally, Self-Instruction (\textit{+ LLM recheck}) aligns with the architecture integrated into our Quality Verification Agent. Therefore, we assert that our approach offers a more thorough solution.

The construction of long-context multi-hop question-and-answer datasets is based on a structured approach leveraging pre-trained document corpora. This section outlines the methodology used for data collection, processing, and validation across multiple domains and languages.

\subsection{Source Data Overview}

The primary source of long-text data is a pre-trained document corpus that spans nine distinct and most widely-used domains~\citep{biancofiore2024interactive}. Inspired by \citet{qin2025survey,zhang2024autocap}, the corpus includes data from the most widely-used bilingual sources (viz. Chinese and English), ensuring a comprehensive multilingual dataset. The domains covered are:
\begin{itemize}[left=2pt,topsep=1pt,itemsep=0pt, parsep=5pt]
    \item \textbf{Books (eBooks)}: A collection of various eBook formats that provide diverse literary content.
    Academic Papers: Scholarly articles sourced from repositories such as arXiv and CNKI. These datasets reflect cutting-edge research across multiple disciplines.
    \item \textbf{Finance}: Data from financial documents and discussions, including the ChatGLM-fin dataset, which encompasses various financial reports and conversational data related to financial analysis.
    \item \textbf{Knowledge}: Information extracted from online encyclopedic sources, including Baike-Wiki and Pile-Wikipedia, covering a broad range of general knowledge.
    \item \textbf{Science}: Data from reputable scientific sources, including Kepuchina and ScienceDaily, that focus on advancements in various scientific fields.
    \item \textbf{Law}: Legal documents and case law from the Pile-Freelaw dataset, providing insight into legal precedents and interpretations.
    \item \textbf{Medicine}: Medical literature, including publications from Pile-PubMed Central, which includes peer-reviewed medical research and case studies.
    \item \textbf{Technology}: Content derived from technical discussions and knowledge-sharing platforms such as Pile-StackExchange.
    \item \textbf{Web Resources}: Web data extracted from open-source platforms, specifically the Pile-OpenWebText2 dataset, reflecting general web-based information.
\end{itemize}

Each domain was selected to ensure the inclusion of diverse, domain-specific content that could support the generation of robust and accurate multi-hop question-and-answer sequences. A more fine-grained analysis can be seen in Figure~\ref{fig:data-analysis} (a).

Additionally, chain-of-thought (CoT) has become a widely-used effective technique for various tasks~\citep{kim-etal-2023-cot,chen-etal-2024-m3cot}. Inspired by this, CoT has the ability to bring powerful performance improvements to the instruction tuning. What's more, as shown in Figure~\ref{fig:case-1}, after adding CoT, the performance of the model has indeed improved, which following the recent reasoning conclusion~\citep{guo2025deepseek,zhuang2023through,chen2024unlocking,chen2025ecm}. Therefore, in all our data synthesis processes, the answer contains a reasoning path.
Furthermore, since LLMs often cannot fit all the document information that is extremely long documents, we perform truncation segmentation on the documents input to the model. After generating the sample, refill the document with other documents to a fixed length.

\subsection{Instruction Dataset Construction}
To expand the domain coverage and handle longer contexts, we extend the instruction fine-tuning data across 9 domains and 2 languages. All base documents are sourced from pre-trained datasets to prevent data leakage. Our Long Multi-hop Instruction-Tuning dataset (\texttt{LongMIT}) results in a retention rate of over 90\% in GPT-4o verification in 200 sampled samples, confirming the high quality and generalizability of our pipeline. To balance the cost and effectiveness of generating data, \texttt{LongMIT} are generated based on Qwen2-72B-Instruct and verified based on InternLM2-20B. We conduct a detailed statistical analysis of sample size and token consumption across various datasets.

\begin{table}[t]
	\centering
    \begin{adjustbox}{width=0.48\textwidth}
        \begin{tabular}{l|cc}
            \toprule
             
            Dataset Name & Sample Size & Token Size \\
            \midrule
            LongAlpaca~\citep{chen2024longlora} & 12.00k & 0.11B \\
            LongAlign~\citep{bai2024long} & 9.89k & 0.17B \\
            ChatQA2~\citep{xu2024chatqa} & 128.00k & 1.22B \\
            NQ~\citep{kwiatkowski2019natural} & 315.20k & 4.56B \\
            \texttt{LongMIT} (Ours) & 64.40k & 5.07B \\
            \bottomrule
        \end{tabular}
    \end{adjustbox}
    \caption{The statistics results of different datasets.\vspace{-4mm}}
    \label{exp:analysis}
\end{table}

Moreover, Table~\ref{exp:analysis} presents the sample and token sizes. Our token count is comparable to that of NQ, but our dataset, while containing fewer samples, outperforms NQ by over 10\%. Notably, despite NQ’s larger sample size, its performance lags behind datasets with fewer samples, such as LongAlpaca and LongAlign.

\subsection{Multi-hop Question and Answer Data Construction}

The construction of multi-hop question-and-answer datasets involved a rigorous process to ensure both linguistic accuracy and domain relevance. The methodology is as follows:

\subsubsection{Dataset Curation} For each domain, data was independently curated to maintain a clear distinction between different knowledge sources. This allows for more focused and accurate multi-hop questions that are relevant to the particular field of study.

\subsubsection{Quality Verification Agent}
\label{append:qva}
The first module in our framework is the Quality Verification Agent, which ensures that the generated questions and answers meet a certain standard of quality. We use InternLM2-20B~\citep{cai2024internlm2} as the backbone and set the quality score threshold to 8.5. Moreover, the prompts are as follows:
\begin{mybox}
    Suppose you are a professional annotator, and you need to annotate the generated questions, rationales, and answers according to the context. Specifically, your tasks are as follows:
    \begin{itemize}[left=2pt,topsep=2pt,itemsep=0pt, parsep=0pt]
        \item First, determine whether the questions and answers are in documents provided in context.
        \item Then, you need to determine whether the problem is a multi-hop problem, using multi-hop logic.
        \item At the same time, you need to judge whether the question conforms to commonsense logic. Does the question conform to common sense in a normal context? Is the logic smooth?
        \item In addition, you need to rate the overall data quality from three aspects: logical rationality and fluency, question complexity, and answer clarity. All scores are between 0 and 10.
        \item Before giving an annotation, you need to give your rationale.
    \end{itemize}

    [[DOCUMENTS]]

    \{chunk\}

    [[QUESTION]]

    \{question\}

    [[ANSWER]]

    \{answer\}

\end{mybox}
\begin{mybox}

\ 

Finally, you should give me an overall quality mark in the format:

```\{"in\_document": BOOL, "domain\_similarity": NUMBER, "quality": NUMBER\}```
\end{mybox}
\subsubsection{Single-hop Question Generation Agent}
\label{append:sqga}
The Single-hop Question Generation Agent is responsible for generating fundamental single-hop questions, which are characterized by their simplicity and directness.

In this framework, we employ Qwen2-72B-Instruct~\citep{yang2024qwen2} as the foundational model, utilizing it to synthesize data through a question-answering paradigm. The process begins with the generation of prompts designed specifically for question creation, initiating a structured approach to the formulation of these queries.
\begin{mybox}
	The document content is as follows:
    
\{chunk\}

Extract the questions contained in the above document, and the extracted questions should meet the following conditions:
\begin{itemize}[left=2pt,topsep=1pt,itemsep=0pt, parsep=0pt]
    \item No pictorial information should be included in the extracted questions;
    \item No referential information should be included in the extracted questions;
    \item Ensure the completeness of the extracted questions; if they are multiple-choice questions, provide corresponding option information, remove line breaks, and place the question body in a single question;

    \item If the document contains concepts such as numbers, time, people, or places, questions that involve this information must be extracted;
    \item The extracted questions should be presented in a parseable list format, such as ["xxx", "xxx"]. If there are no valuable questions, output an empty list [];
    \item Try to extract as many valuable questions as possible, but do not include duplicate questions;
    \item Extract no more than three questions;
\end{itemize}

Extracted questions:
\end{mybox}

Based on the questions extracted, the prompt for answer generation is as follows:
\begin{mybox}
	Generate answers to a given series of questions based on the content of the document, which must meet the following conditions:
    \begin{itemize}[left=2pt,topsep=1pt,itemsep=0pt, parsep=0pt]
        \item Respond based on the content in the document;
        \item If there is no corresponding answer to the question in the document, please reply based on your own knowledge;
        \item If the question is about factual issues such as numbers, time, people, places, etc., please provide the answer directly, and different question and answer pairs should be distinguished by line breaks;
    \end{itemize}

The document content is as follows:

\{chunk\}

The problems are as follows:

\{question\}

The corresponding answers are as follows:
\end{mybox}

\subsubsection{Multiple Question Sampling}

This strategy further enhances the generation of multi-hop instructions by selecting questions that address diverse elements within the document. This method facilitates the creation of comprehensive, multi-hop, long-text question-answer datasets that are meticulously customized to reflect the characteristics and requirements of specific domain data sources.
The organization of the relevant documents begins by embedding them into vectors, where BGE-zh-1.5 and BGE-en-1.5~\citep{bge_embedding} models are used to map the documents into 768-dimensional vectors. Following the methods inspired by \citet{shi2024incontext}, the document vectors are embedded using Faiss to facilitate storage and efficient retrieval. This process relies on measuring vector distances to retrieve the 10 nearest documents for each document, creating a document graph.

Subsequently, a circular search strategy is employed to generate paths that consist of multiple documents, with the maximum path length constrained to 20. This process continues until all documents are sampled, with these paths serving as the initial sets of multiple related documents.

After conducting a sampling analysis, we observed the hop distribution in the constructed data, as illustrated in Figure~\ref{fig:data-analysis} (b). Additionally, the distribution corresponding to the sampling strategy is depicted in Figure~\ref{fig:data-analysis} (c).

\subsubsection{Multi-hop Question Merging Agent}
Multi-hop questions are designed to require reasoning across multiple data points, either within a single domain or spanning different domains. This approach ensures that responses cannot be derived from isolated facts; rather, they necessitate a more profound comprehension and integration of the dataset's overall content.

To achieve this, the Multi-hop Question Merging Agent consolidates single-hop questions into well-structured multi-hop queries. This process demands information synthesis from various sections of the document, promoting a deeper level of understanding and engagement. For the model architecture, we employ Qwen2-72B-Instruct~\citep{yang2024qwen2} as the base model. The specific prompt for merging two QA pairs is as follows:

\begin{mybox}
    Based on the given two question-answer pairs, synthesize up to one question-answer pair that matches the real scenario. The synthesized question-answer pair should meet the following conditions:

    \begin{itemize}[left=2pt,topsep=1pt,itemsep=0pt, parsep=0pt]
        \item If both questions and answers are time-related, a comparative question can be synthesized to compare the order in which two events occur;
        \item If both questions and answers are related to the character, it can be synthesized to determine which character better fits the description of the composite question;
        \item The synthesized answer should provide the corresponding reasoning process, and the synthesized answer should make as much use of the content in the given two answers as possible;
        \item Do not arbitrarily change the original information of two questions and answers;
        \item The generated questions and answers are strictly output in JSON format using \{"question": xxx, "answer": xxx\}. Synthesized question-answer pairs should not have any line breaks;
    \end{itemize}
\end{mybox}
\begin{mybox}

The correct answers to two questions are as follows:

\{qa1\}

\{qa2\}

The synthesized question-answer pair is:
\end{mybox}

\begin{table*}[t]
	\centering
    \begin{adjustbox}{width=0.90\textwidth}
        \begin{tabular}{lcccccccccc}
            \toprule
             & \multicolumn{5}{c}{InterLM-2.5-7B-Enhance} & \multicolumn{5}{c}{InterLM-2.5-7B-Enhance + \texttt{LongMIT}} \\
            \cmidrule(lr){2-6}
            \cmidrule(lr){7-11}
             & 4k & 8k & 16k & 32k & 128k & 4k & 8k & 16k & 32k & 128k \\
            \midrule
            S-NIAH Subtask-1 & 99.00  & 99.00  & 100.00  & 100.00  & 16.00  & 99.00  & 99.00  & 99.00  & 99.00  & 97.00 \\
            S-NIAH Subtask-2 & 100.00  & 99.00  & 100.00  & 99.00  & 97.00  & 100.00  & 100.00  & 100.00  & 100.00  & 100.00 \\
            S-NIAH Subtask-3 & 99.00  & 98.00  & 99.00  & 99.00  & 100.00  & 99.00  & 99.00  & 99.00  & 99.00  & 100.00 \\
            MK-NIAH Subtask-1 & 97.00  & 98.00  & 97.00  & 88.00  & 58.00  & 100.00  & 100.00  & 98.00  & 99.00  & 90.00 \\
            MK-NIAH Subtask-2 & 99.00  & 99.00  & 96.00  & 81.00  & 28.00  & 99.00  & 100.00  & 100.00  & 95.00  & 63.00 \\
            MK-NIAH Subtask-3 & 95.00  & 90.00  & 56.00  & 14.00  & 0.00  & 96.00  & 91.00  & 70.00  & 33.00  & 2.00 \\
            MV-NIAH & 99.25  & 99.50  & 99.50  & 94.50  & 84.50  & 99.00  & 99.00  & 97.00  & 93.50  & 89.50 \\
            MQ-NIAH & 98.00  & 98.75  & 97.50  & 94.00  & 86.00  & 100.00  & 100.00  & 100.00  & 99.25  & 94.25 \\
            VT & 91.20  & 91.80  & 98.60  & 97.40  & 0.00  & 96.60  & 97.80  & 98.80  & 95.60  & 94.20 \\
            FWE & 85.33  & 87.00  & 84.67  & 91.00  & 71.67  & 86.00  & 89.00  & 86.67  & 90.67  & 78.33 \\
            CWE & 83.40  & 67.10  & 34.70  & 26.40  & 0.10  & 75.60  & 40.40  & 8.50  & 6.10  & 0.20 \\
            Single Hop QA & 90.00  & 80.00  & 81.00  & 75.00  & 42.00  & 92.00  & 84.00  & 82.00  & 80.00  & 58.00 \\
            Multi Hop QA & 70.00  & 67.00  & 64.00  & 53.00  & 35.00  & 73.00  & 72.00  & 68.00  & 63.00  & 47.00 \\
            \midrule
            Average & 92.78  & 90.32  & 85.23  & 77.87  & 47.56  & \textbf{93.48}  & \textbf{90.09}  & \textbf{85.15}  & \textbf{81.01}  & \textbf{70.27} \\
            \bottomrule
        \end{tabular}
    \end{adjustbox}
    \caption{The evaluation performance on Ruler~\citep{hsieh2024ruler} benchmark based on \texttt{LongMIT} extended to 128K.\vspace{-4mm}}
    \label{exp:ruler-exp}
\end{table*}

\section{Highest Quality Strategy Details}
\label{append:best-precision}
\subsection{Different Generation Strategy}
\subsubsection{Generation Strategy Defination}
\textbf{LongMIT+Best-Strategy} means that add all strategies that yield better performance but may incur higher costs. These include using GPT-4o as the backbone model, incorporating additional rationales, merging questions with corresponding documents, and other related techniques. More details are described in Sec.~\ref{append:best-precision-imp}.

\noindent\textbf{Self-Instruct strategy} involves prompting GPT-4o to autonomously generate questions and corresponding answers based on the provided document, leveraging its self-instruction capabilities for required outputs.
\subsubsection{Generation Strategy Discussion}
As shown in Figure~\ref{fig:overall-cost}, the quality of the data synthesized using Qwen2 within the \texttt{MIMG} framework significantly exceeds that generated by GPT-4 using the Self-Instruct strategy. This illustrates the effectiveness of our framework, even more effective than backbone replacement. Furthermore, as illustrated in Table~\ref{exp:main-exp}, our approach consistently outperforms prior efforts utilizing GPT-3.5, GPT-4, and even human-created data, which demonstrates and further proves the effectiveness of our framework.

In addition, GPT4 is often impractical for training data synthesis. It is also important to note that the large-scale training data, distilled from only 4 billion tokens using GPT-4, incurs a cost exceeding \$10,000, rendering such an approach impractical for many applications.

\begin{figure}[t]
	\centering
	\begin{minipage}{0.46\textwidth}
		\centering
		\includegraphics[width=0.99\linewidth]{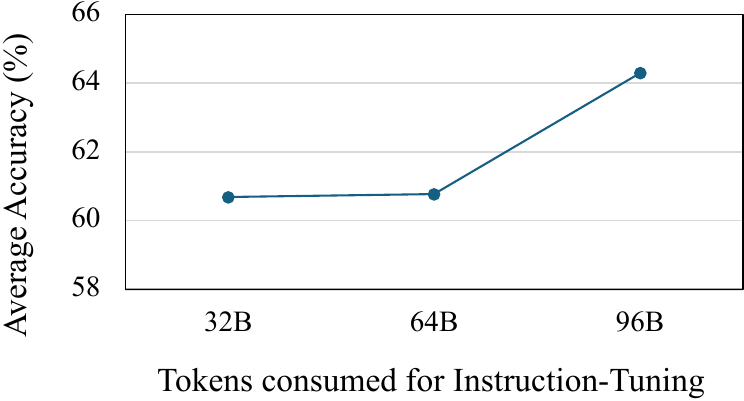}
		\captionof{figure}{Analysis of the impact of different training dataset sizes on the average accuracy score.
		}
		\label{fig:data-size-affect}
	\end{minipage}
	\hfill
	\begin{minipage}{0.49\textwidth}
		\centering
		\includegraphics[width=0.99\linewidth]{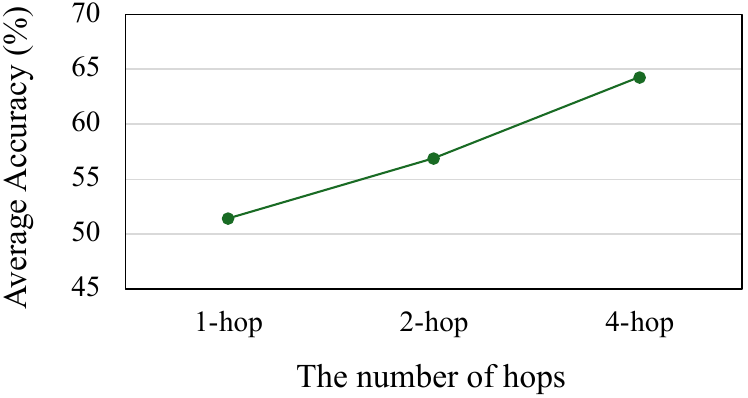}
		
		\captionof{figure}{Analysis of the impact of hop on model performance, where 1-hop is the reproduced version of the Quest~\citep{gao2024quest} dataset.
		}
		\label{fig:hop-affect}
	\end{minipage}
\end{figure}

\subsection{Implementation Details}
\label{append:best-precision-imp}
To achieve the highest quality data, we deliberately prioritize the use of GPT-4o as the backbone for all processes, fully disregarding cost constraints. This decision is driven by the understanding that ensuring the best data quality is paramount for the success of our project. Furthermore, to maintain and enhance performance during the exploration phase, we implement a range of strategies aimed at maximizing the data retention rate.

Specifically, for the Quality Verification Agent, we employ a multi-faceted approach that includes more-perspectives scoring mechanisms, the addition of rationales, the integration of multiple perspectives, and the application of detailed guidelines.
For the Single-hop Question Generation Agent, we have adopted a question-then-answer strategy. This approach is complemented by the incorporation of rationales, which provide context and justification for each query generated.
Additionally, we require LLMs to generate only one question per query, which is intended to reduce the logical burden on the model, thereby improving the coherence and relevance of the questions produced.
In the case of Multiple Question Sampling, we utilize BGE embeddings for the retrieval of questions. This technique is applied both within individual documents (intra-document) and across multiple documents (inter-document).
Finally, for the Multi-hop Question Merging Agent, we employ a strategy that involves merging questions and answers using document references. This method ensures that the merged questions and answers are contextually aligned and coherent. Notably, we have opted to remove the rationale for merging in this process, as we found that it adds unnecessary complexity without significantly improving the quality of the merged content.

\begin{table}[t]
	\centering
    \begin{adjustbox}{width=0.48\textwidth}
        \begin{tabular}{lccc}
            \toprule
             & High-quality & Diversity & Multi-hop \\
            \midrule
            \texttt{Self-Instruct} & 61.3 & 53.4 & 33.1 \\
            \midrule
            \texttt{MIMG} & 94.8 & 88.2 & 94.8 \\
            \ \ \ \ \texttt{w/o MQMA} & 85.0 & 85.7 & 45.7\\
            \ \ \ \ \texttt{w/o MQS} & 92.8 & 65.4 & 89.4\\
            \ \ \ \ \texttt{w/o SQGA} & 82.4 & 69.4 & 73.2\\
            \ \ \ \ \texttt{w/o QVA} & 78.3 & 88.1 & 91.0\\
            \bottomrule
        \end{tabular}
    \end{adjustbox}
    \caption{Results of the ablation study on MIMG, reporting the data quality score as the average of three independent manual evaluations.\vspace{-4mm}}
    \label{exp:ablation}
\end{table}

\begin{table*}[t]
  \centering
  \begin{adjustbox}{width=0.92\textwidth}
    \begin{tabular}{lcccc}
      \toprule
      Model & inst\_level\_loose\_acc & inst\_level\_strict\_acc & prompt\_level\_loose\_acc & prompt\_level\_strict\_acc \\
      \midrule
      Llama-3-8B-ProLong & 0.1259 & 0.1199 & 0.0610 & 0.0536 \\
      \ \ +ChatQA2            & 0.0156 & 0.0132 & 0.0074 & 0.0037 \\
      \ \ +LongMIT            & 0.1583 & 0.1463 & 0.0795 & 0.0684 \\
      \bottomrule
    \end{tabular}
  \end{adjustbox}
  \caption{The instruction-following capabilities on IFEval for different instruction datasets.}
  \label{tab:ifeval}
\end{table*}

\begin{table}[t]
  \centering
  \begin{adjustbox}{width=0.92\linewidth}
    \begin{tabular}{lccc}
      \toprule
       & High-quality & Diversity & Multi-hop \\
      \midrule
      \rowcolor{gray!8}\multicolumn{4}{c}{InternLM2-Chat-20B~\citep{cai2024internlm2}}\\
      \midrule
      \texttt{Self-Instruct}   & 51.0    & 44.0    & 52.0    \\
       \texttt{MIMG}            & 86.0    & 66.0    & 80.0    \\
      \midrule
      \rowcolor{gray!8}\multicolumn{4}{c}{Qwen-72B-Instruct~\citep{yang2024qwen2}}\\
      \midrule
      \texttt{Self-Instruct}   & 61.3  & 53.4  & 33.1  \\
      \texttt{MIMG}  & 94.8  & 88.2  & 94.8  \\
      \midrule
      \rowcolor{gray!8}\multicolumn{4}{c}{GPT-4o~\citep{achiam2023gpt}}\\
      \midrule                            
      \texttt{Self-Instruct}   & 86.0    & 81.0    & 66.0    \\
      \bottomrule
    \end{tabular}
  \end{adjustbox}
  \caption{The human-annotated quality score for different LLMs and strategies.}
  \label{tab:quality_scores}
\end{table}

\subsection{Ablation Analysis}
We analyze the contributions of various agent components to the quality of human-annotated data. As summarized in Table~\ref{exp:ablation}, each component positively impacts the overall quality. The Multi-hop Question Merger Agent notably improves the multi-hop quality, while the Multiple Question Sampling mechanism increases data diversity. The Single-hop Question Generation Agent is crucial for enhancing both quality and diversity. Lastly, the Quality Verification Agent acts as a safeguard, ensuring a lower bound of model quality and further improving data integrity.

\subsection{More Backbone Exploration}
To assess the data generation quality across additional backbone architectures, we manually labeled 100 samples for both open-source and closed-source models. As reported in Table~\ref{tab:quality_scores}, integrating \texttt{MIMG} with \texttt{InternLM2} yields a marked improvement in output quality. Moreover, \texttt{LongMIT} surpasses the combination of \texttt{GPT-4o} and \texttt{Self-Instruct} on our quality metrics. We will elaborate on these findings in the next version.

\subsection{Instruction Capabilities}
As \texttt{MIMG} is designed to enhance models’ ability to follow instructions over extended contexts. Accordingly, it is essential both to improve instruction-following performance and to preserve this ability in long-context settings. Because there is currently no benchmark dedicated to long-context instruction following, we evaluated instruction adherence using two established short-context benchmarks: \texttt{IFEval} and \texttt{ArenaHard}.

As shown in Table~\ref{tab:ifeval}, remarkably, the model trained with \texttt{LongMIT} not only achieved substantial gains in handling long contexts but also outperformed the base model on the \texttt{IFEval} benchmark. In contrast, training on \texttt{ChatQA2} led to a pronounced decline in instruction-following performance on \texttt{IFEval}.

In addition, as presented in Table~\ref{tab:arena_hard_results}, on the more demanding short-context instruction-following benchmark \texttt{ArenaHard}, our method continues to outperform \texttt{ChatQA2} by a substantial margin. Moreover, the improvements to long-context processing do not compromise its effectiveness on challenging instruction-following tasks.

\begin{table}[t]
  \centering
  \begin{adjustbox}{width=\linewidth}
    \begin{tabular}{l c c}
      \toprule
      \texttt{ArenaHard}                  & Score & 95\% Confidence Interval \\
      \midrule
      \texttt{Llama-3-8B-ProLong}         & 7.2   & $(-1.2,\;1.2)$           \\
      \ \ \ \ \texttt{+ChatQA2}                   & 4.4   & $(-0.9,\;0.8)$           \\
      \ \ \ \ \texttt{+LongMIT}                   & 6.7   & $(-1.1,\;1.0)$           \\
      \bottomrule
    \end{tabular}
  \end{adjustbox}
  \caption{The instruction-following capabilities on ArenaHard for different instruction datasets.}
  \label{tab:arena_hard_results}
\end{table}

\begin{figure*}[t]
	\centering
	\includegraphics[width=0.90\textwidth]{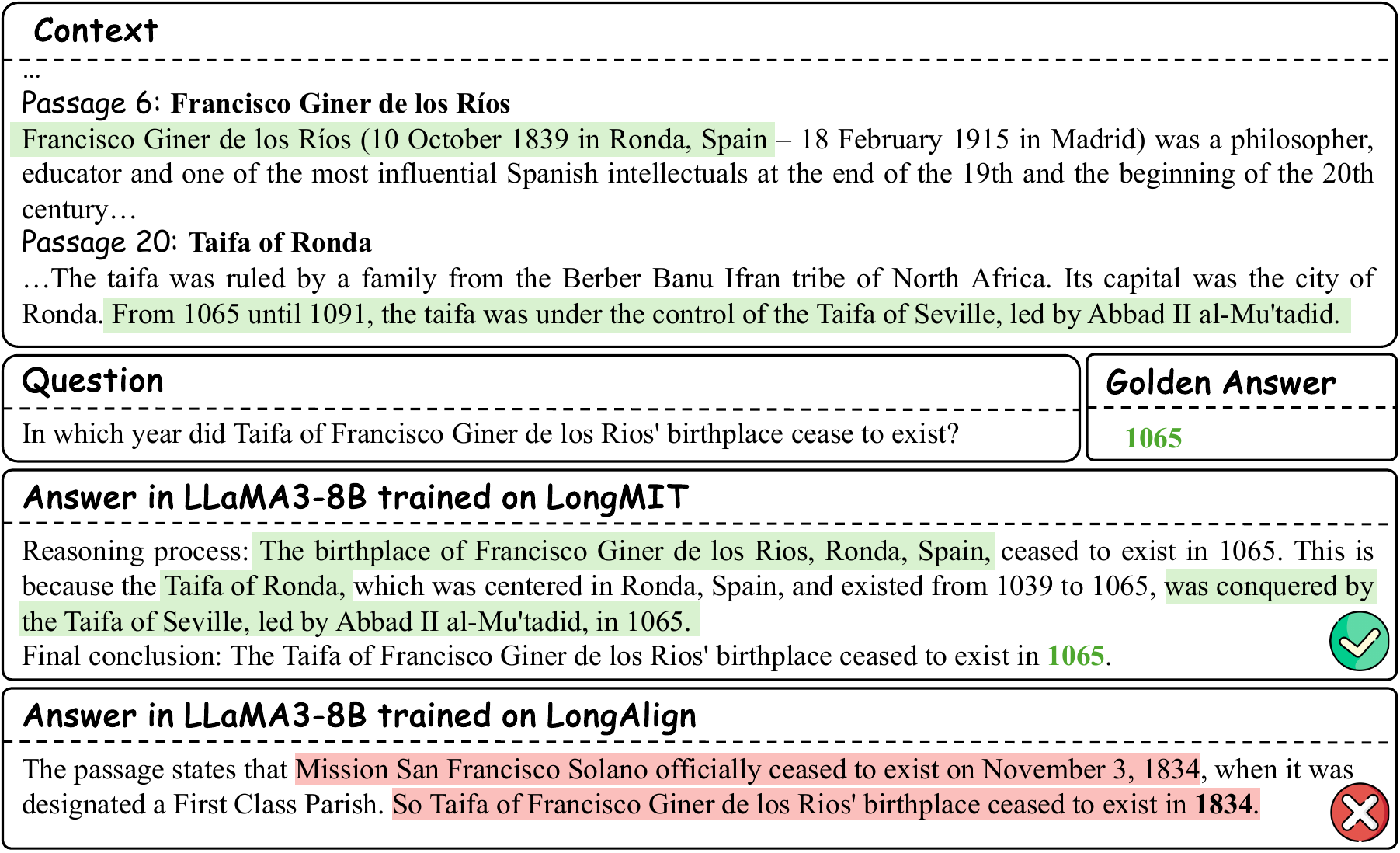}
	\caption{The case study of the effectiveness of \texttt{LongMIT}.
	}
	\label{fig:case-2}
\end{figure*}

\section{Instruction Tuning Experiments Details}
\label{append:train}

\subsection{Training Details}
All models were trained using 64 A800*80G GPUs with the DeepSpeed+ZeRO-1 framework. The maximum sequence length was set from 4K to 128K, with any sequences exceeding this length truncated from the right. The training process utilized the Adam optimizer with a learning rate of $3 \times 10^{-5}$, $\beta_1 = 0.9$, and $\beta_2 = 0.95$.

To enhance training efficiency, we employed a packing strategy that concatenates training samples to reach the maximum sequence length. Additionally, Flash Attention~\citep{NEURIPS2022_67d57c32,dao2024flashattention} is used to accelerate the computation of the attention mechanism. The global batch size consisted of 4 million tokens, and the entire dataset is trained over one epoch.

\subsection{Evaluation Details}
Based on the methodology proposed by \citet{bai2024long}, evaluating Token F1 using a model optimized through Chain of Thought (CoT)~\citep{NEURIPS2022_9d560961} reasoning proves to be challenging~\citep{chen2025towards}. To address this limitation, we employ GPT-4 as a consistency evaluator. Our testing demonstrates that the error rate of GPT-4 in this role remains consistently low, with deviations falling within a 2\% margin. The corresponding prompt used is outlined below:
\begin{mybox}
    Suppose you are a professional annotator. Given the result predicted by a model, you  need to annotate whether the ``[[PREDICTION]]`` is consistent with the given ``[[REFERENCE]]`` based on the ``[[QUESTION]]``.

[[QUESTION]]

\{question\}

[[PREDICTION]]

\{predictions\}

[[REFERENCE]]

\{answer\}

Finally, you should give me an annotation in the format:

```
\{
    "short\_pred\_answer": "xxx", 
    "predict\_consistency": BOOL
\}
```
\end{mybox}

\begin{figure*}[t]
	\centering
	\includegraphics[width=0.90\textwidth]{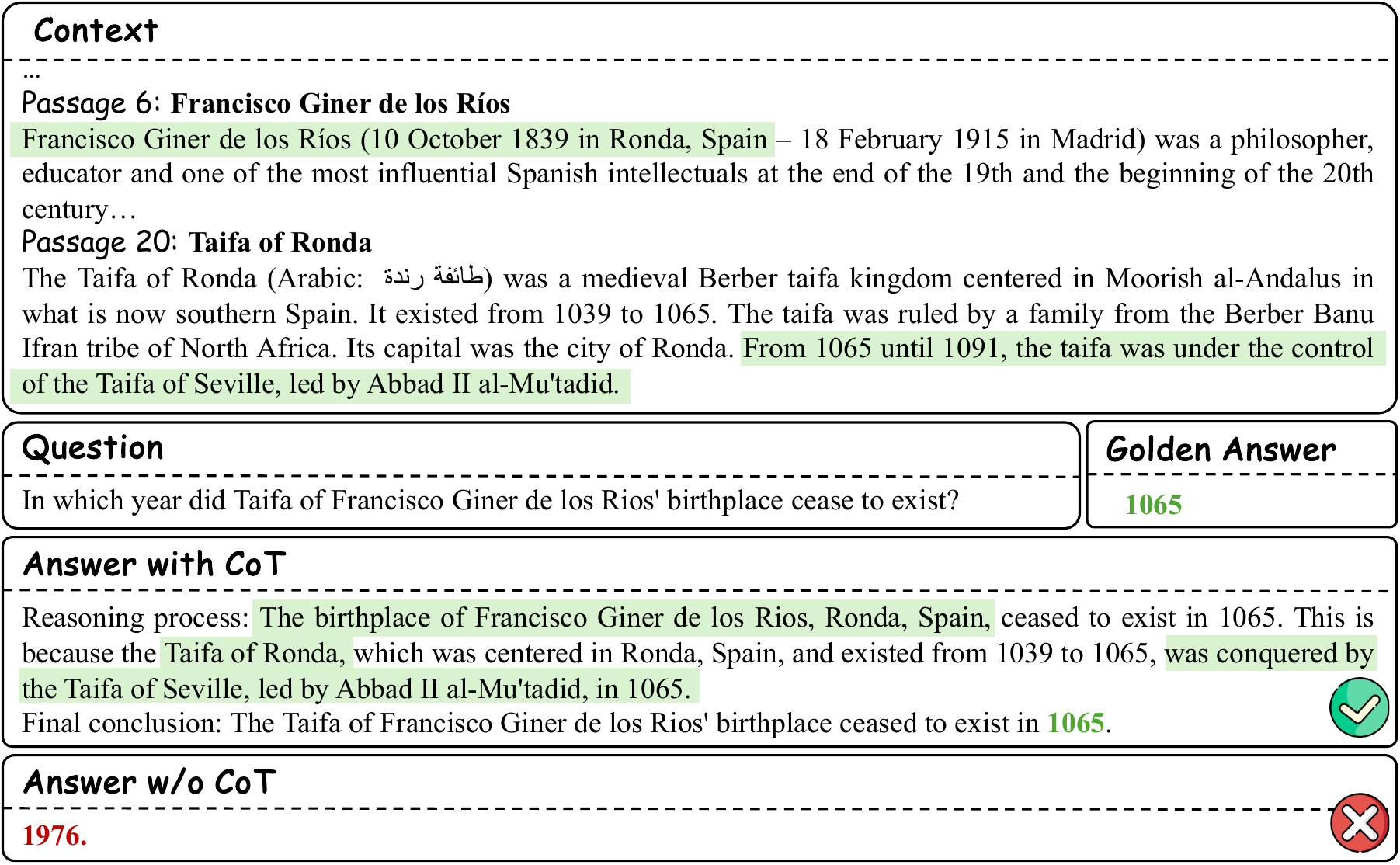}
	\caption{The case study of whether utilize reasoning process for instruction tuning.
	}
	\label{fig:case-1}
\end{figure*}

\section{Case study}
\label{append:case}
To gain a more nuanced and intuitive qualitative understanding of our model's performance, we conducted a detailed case study, resulting in two significant findings:
\begin{itemize}[left=2pt,topsep=1pt,itemsep=0pt, parsep=0pt]
    \item \textbf{Impact of Instruction Quality:} As illustrated in Figure~\ref{fig:case-1}, models trained with high-quality multi-hop instruction data, specifically the \texttt{LongMIT} dataset, exhibit enhanced logical reasoning capabilities. These models are better equipped to process and analyze extensive textual information, enabling them to derive more accurate and reliable reasoning. In contrast, models trained using traditional, lower-quality instruction data, such as \texttt{LongAlign}~\citep{bai2024long}, demonstrate a reduced capacity for logical reasoning. This comparison underscores the importance of the quality of training data in developing models that can effectively handle complex reasoning tasks, especially when dealing with long and intricate texts.
    \item \textbf{Role of Rationale Incorporation in Training:} Furthermore, as depicted in Figure~\ref{fig:case-2}, our analysis reveals that the inclusion of additional rationales during the training process significantly enhances the model's ability to focus on relevant information within long texts and make precise inferences. This finding is particularly evident when comparing models that underwent Chain-of-Thought (CoT)~\citep{NEURIPS2022_9d560961} training with those that did not. Specifically, models that lacked CoT training tend to falter during inference, often generating erroneous outputs, such as the completely incorrect answer "1976". On the other hand, models that were fine-tuned with CoT training not only demonstrate a coherent logical reasoning process but also consistently arrive at the correct answer, "1065". This result highlights the critical role of rationale-based training in improving the model's reasoning accuracy and its ability to tackle complex inferential challenges.
\end{itemize}

\section{Discussion about Long Code Data Generation}
\label{append:code}
In both coding and mathematical settings, it is essential to perform multi-step reasoning across source files and scholarly articles. To investigate this requirement, we evaluate our framework on a set of coding tasks. The resulting code–question–answer dataset exhibited a high level of quality, with approximately 80\% of entries rated as satisfactory. Furthermore, by consolidating related questions, we are capable to prompt the advanced LLM (GPT-4o) to generate code-generation queries of competition caliber.

Below, we illustrate a simple example in which the context consists of code documents from the ``torch.nn'' module. First, the Single-Hop Question Generation Agent generates single-hop question-answer pairs as follows:
\begin{mybox}
\textbf{Question 1:} What is the purpose of nn.LeakyReLU?

\textbf{Answer 1:} nn.LeakyReLU is similar to nn.ReLU but allows a small, non-zero slope for negative values, controlled by a "negative\_slope" parameter, making it less likely for neurons to become inactive.

\textbf{Question 2:} What is the nn.ReLU?

\textbf{Answer 2:} nn.ReLU (Rectified Linear Unit) applies the rectified linear unit function element-wise, where any negative input values are set to zero, leaving positive values unchanged.
\end{mybox}
After that, Multi-hop Question Merging Agent generates the merged question-answering pair as:
\begin{mybox}
\textbf{Question:} How does the behavior of nn.ReLU differ from nn.LeakyReLU in terms of their activation functions when used in a neural network constructed with nn.Sequential?

\textbf{Answer:} In a neural network constructed with nn.Sequential, nn.ReLU applies the rectified linear unit function element-wise, where negative input values are set to zero, while positive values remain unchanged.
\end{mybox}
\begin{mybox}
In contrast, nn.LeakyReLU behaves similarly but allows a small, non-zero slope for negative values through a 'negative\_slope' parameter, which helps prevent neurons from becoming inactive. This means nn.LeakyReLU can potentially maintain more active neurons compared to nn.ReLU in sequential layers.
\end{mybox}
Further, We try to propose a new prompt for Multi-hop Question Merging Agent to generate more difficult questions for competitive code based on merged question.
\begin{mybox}
Using the following question and answer related knowledge and meanings as a reference, generate additional programming challenges and their solutions specifically designed for high-level coding competitions, such as Olympiad-style problems or advanced AI-related tasks (like those on Kaggle). The focus should be on tasks that require efficient code generation, algorithm design, or AI model implementation.
\end{mybox}
As shown below, large language models (LLMs) can efficiently produce sophisticated code‐generation datasets. This capability represents a compelling and promising research direction.
One of generated questions is presented as follows:
\begin{mybox}
\textbf{Problem Statement:}

You are tasked with using a genetic algorithm (GA) to optimize the hyperparameters of a neural network for a given task (e.g., image classification using CIFAR-10 dataset). You need to optimize parameters such as the number of layers, the number of neurons in each layer, and the activation function used at each layer.

\textbf{Input:}

A dataset such as CIFAR-10 or MNIST (for image classification).
An initial population of neural network configurations represented by a list of hyperparameters:
\begin{itemize}[left=2pt,topsep=1pt,itemsep=0pt, parsep=0pt]
    \item Number of layers.
    \item Number of neurons per layer.
\end{itemize}
\end{mybox}
\begin{mybox}
\begin{itemize}[left=2pt,topsep=1pt,itemsep=0pt, parsep=0pt]
    \item Activation functions (ReLU, LeakyReLU, Sigmoid, etc.).
\end{itemize}
\end{mybox}

Further, the corresponding generated answer is shown below:
\begin{mybox}
\textbf{Output:}

The optimized neural network configuration after a set number of generations.

\textbf{Solution:}
\end{mybox}
\begin{lstlisting}[language=Python]
import random
from sklearn.datasets import fetch_openml
from sklearn.model_selection import train_test_split
import torch
import torch.nn as nn
import torch.optim as optim
import torch.nn.functional as F

# Simple feedforward neural network
class NeuralNet(nn.Module):
    def __init__(self, layers, activation_fn):
        super(NeuralNet, self).__init__()
        self.layers = nn.ModuleList()
        for i in range(len(layers) - 1):
            self.layers.append(nn.Linear(layers[i], layers[i+1]))
        self.activation_fn = activation_fn

    def forward(self, x):
        for layer in self.layers:
            x = layer(x)
            x = self.activation_fn(x)
        return x

# Genetic algorithm for hyperparameter optimization

def genetic_algorithm(dataset, generations=10, population_size=10):
    population = []
    for _ in range(population_size):
        layers = [random.randint(5, 100) for _ in range(random.randint(3, 6))]  # Number of neurons in layers
        activation_fn = random.choice([nn.ReLU(), nn.LeakyReLU(), nn.Sigmoid()])
        population.append((layers, activation_fn))

    best_score = float('inf')
    best_model = None

    for generation in range(generations):
\end{lstlisting}

\ 
\begin{lstlisting}[language=Python]
        for layers, activation_fn in population:
            # Convert layers to torch model

            model = NeuralNet(layers, activation_fn)
            optimizer = optim.Adam(model.parameters())
            criterion = nn.CrossEntropyLoss()

            # Training loop on dataset
            model.train()
            optimizer.zero_grad()
            # Assume a dummy data loading and training process here
            loss = criterion(model(dummy_inputs), dummy_labels)
            loss.backward()
            optimizer.step()

            if loss.item() < best_score:
                best_score = loss.item()
                best_model = model

        # Selection for next generation (randomly choose top models for simplicity)
        population = random.sample(population, k=population_size)

    return best_model

# Example usage
best_model = genetic_algorithm(fetch_openml("CIFAR_10"))
\end{lstlisting}

\label{sec:appendix}

\end{document}